%% file: 00_main.tex
\documentclass[anonymous=false,acmsmall,screen,nonacm]{acmart}
\usepackage{gensymb}
\usepackage[subtle]{savetrees}
\usepackage{etoolbox}
\usepackage{graphicx}
\usepackage{subcaption}
\usepackage[export]{adjustbox}
\usepackage{wrapfig}
\makeatletter
\patchcmd{\@verbatim}
  {\verbatim@font}
  {\verbatim@font\small}
  {}{}
\makeatother

\AtBeginDocument{%
  \providecommand\BibTeX{{%
    \normalfont B\kern-0.5em{\scshape i\kern-0.25em b}\kern-0.8em\TeX}}}

\usepackage{multirow}
\usepackage{xcolor} 
\usepackage{soul}
\begin{document}

\title[Making Sense of Robots in Public Spaces]{Making Sense of Robots in Public Spaces:\\A Study of Trash Barrel Robots}

\author{Fanjun Bu}
\email{fb266@cornell.edu}
\orcid{0000-0002-9953-7347}
\affiliation{%
  \institution{Cornell University, Cornell Tech}
  \streetaddress{2 W Loop Rd}
  \city{New York}
  \state{New York}
  \country{USA}
  \postcode{10044}
}

\author{Kerstin Fischer}
\orcid{0000-0003-1987-5344}
\affiliation{%
  \institution{Southern Denmark University}
  \country{Denmark}
 }
\email{kerstin@sdu.dk}

\author{Wendy Ju}
\email{wendyju@cornell.edu}
\orcid{0000-0002-3119-611X}
\affiliation{
    \institution{Jacobs Technion-Cornell Institute, Cornell Tech}
    \state{New York}
    \country{USA}
}

\renewcommand{\shortauthors}{Fanjun Bu, Kerstin Fischer, \& Wendy Ju}

\begin{abstract}
In this work, we analyze video data and interviews from a public deployment of two trash barrel robots in a large public space to better understand the sensemaking activities people perform when they encounter robots in public spaces. Based on an analysis of 274 human-robot interactions and interviews with \textit{N=}65 individuals or groups, we discovered that people were responding not only to the robots or their behavior, but also to the general idea of deploying robots as trashcans, and the larger social implications of that idea. They wanted to understand details about the deployment because having that knowledge would change how they interact with the robot. Based on our data and analysis, we have provided implications for design that may be topics for future human-robot design researchers who are exploring robots for public space deployment. Furthermore, our work offers a practical example of analyzing field data to make sense of robots in public spaces.
\end{abstract}

\begin{teaserfigure}
\centering
  \includegraphics[width=0.95\textwidth]{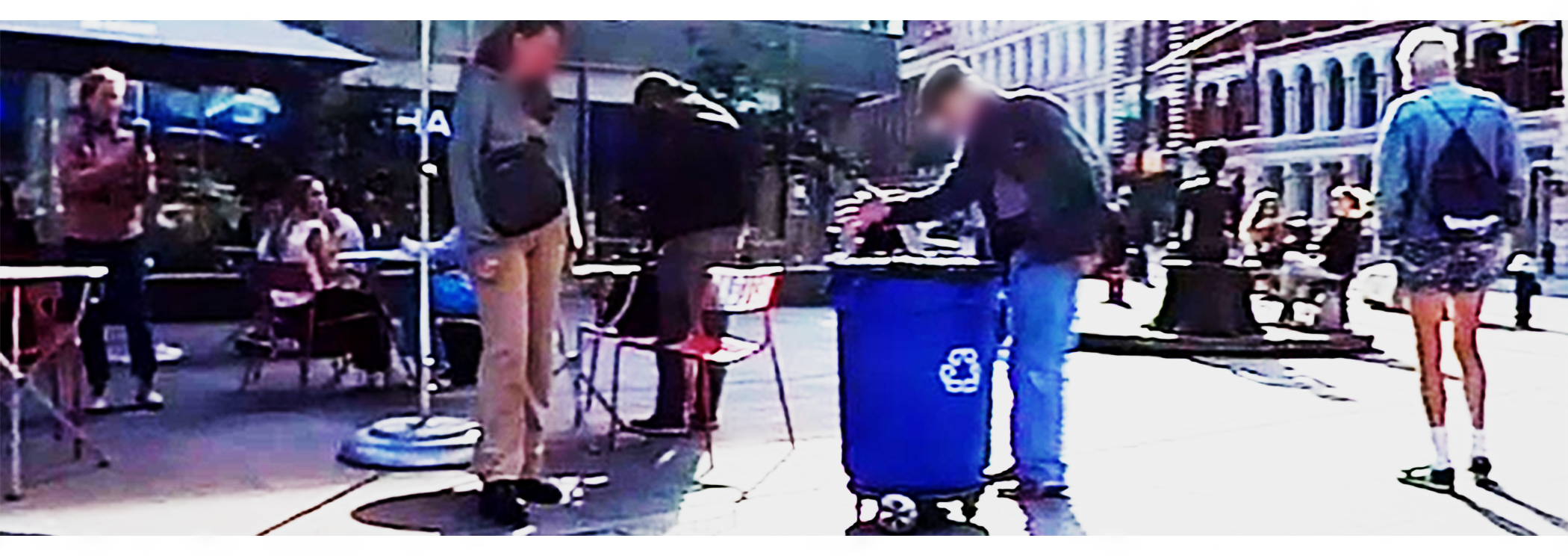}
  \caption{People in public space trying to make sense of the trash barrel robots.}
  \Description{}
  \label{fig:teaser}
\end{teaserfigure}


\maketitle

\input{01_introduction}
\input{02_relatedwork}
\input{03_method}
\input{04_analysis}
\input{05_discussion}
\input{06_limitation}
\input{07_conclusion}

\begin{acks}
This research was conducted under Cornell Tech’s IRB Protocol \#IRB0147271, with sponsorship from Tata Consultancy Services and the Jacobs Urban Tech Hub at Cornell Tech, as well as support from NSF \#2423127 FRR: Social and Contextual Models of Interaction with Everyday Robots. We appreciate the wizards who played integral roles behind the scenes: Rei Lee, Saki Suzuki, Ricardo Gonzalez, Stacey Li, Maria Teresa Parreira, and Jorge Pardo Gaytan.
\end{acks}

\bibliographystyle{ACM-Reference-Format}
\bibliography{hri-sensemaking}

\appendix
\section{Appendix}
\subsection{Wizard Instructions}
\label{wizard_instructions}
\input{wizard_instruction}

\subsection{Interview Questions}
\label{questions}
\input{interview_questions}
\subsection{Authorization and Consent}
\label{consent}
\input{authorization_consent}
\end{document}

%% file: 01_introduction.tex
\section{Introduction}
How should we design robots to be deployed in public spaces? From more than two decades of research in human-robot interaction, we have a good idea of how people make sense of robots in one-on-one lab encounters. For instance, we know that people tend to interact with robots as if they were social agents (e.g. \cite{clark_fischer_2023}, among many others), and much research in robot (behavior) design addresses how such interactions can be facilitated.

Much less is known about robots in the wild \cite{jung2018robots}; people engage in the sensemaking process when novel technologies are introduced to their spaces \cite{forlizzi2007robotic}, but so far it is not clear how this sensemaking differs from lab contexts and what design implications emerge from that. Consequently, in this paper, we ask how we can design robots for public spaces to address the sensemaking processes that occur in the field, whereby sensemaking we mean observable traces of how people make sense of the robots and what they are doing in the public space, what actions people carry out to decide how to interact with a robot -- if at all -- as well as the cognitive and cultural resources that they evoke in understanding what the robots are up to.

Specifically, we investigate the deployment of two trash barrel robots in Astor Place, Manhattan, New York City. We focus on how the general public reacted to the introduction of service robots in public spaces, what resources they draw on, and how they make sense of the robots' presence. The analysis of both post-interaction interviews and video footage reveals that in a public space like Astor Place, people draw on a very broad range of resources, making use of cultural knowledge, knowledge of human nature, communal common ground, and personal experience from interactions with the robots to make sense of the robot deployment, taking into account not only the robots themselves but also their operators, possible deployers, the city, vulnerable co-citizens and society at large when interpreting the novel situation, to mention just a few resources that contribute to the sensemaking process. These results have consequences for how robots should be designed for deployment in public spaces. Our analysis also contributes to the methodology of sensemaking analysis through in-field deployments, serving as an example for future studies and practices in understanding human-robot interactions in real-world contexts.

%% file: 02_relatedwork.tex
\section{Related Work}

 When people run into robots, they need to turn the encountered circumstances into situations where they can act upon, a process called sensemaking \cite{weick2005organizing}. The concepts involved in sensemaking research are informed by research from a broad range of disciplines, such as HCI, systems engineering, knowledge management, organizational communication, and user studies \cite{turner2023multifaceted}. Here we discuss sensemaking as a possible method for providing designerly ways \cite{cross2006designerly, lupetti2021designerly} to address the emergent design interactions people will have with robots in their public spaces. 
 
\subsection{Public deployments of robots}

While many researchers in HRI have focused on conducting robot studies "in the wild," their arguments for these often focus on the public environments as being the "ultimate test" for robots---to understand how they will be used \cite{sabanovic2006robots}, what people consider to be normal or breaches of expectation \cite{weiss2008methodological}, or what features will lead people to engage with robots more \cite{moshkina2014social}, outside of the confines of the lab.

Beyond the 'ultimate test'-view, roboticist Pericle Salvini offers an "urban robotics perspective," which suggests that the focus of such studies should be on the "emerging properties" that result from the in-situ interactions \cite{salvini2018urban}. Instead of testing hypotheses,  profiling user population behaviors, or identifying qualities of robots fit or unfit for the urban environment, as has been the case in HRI research to date, the urban robotics perspective centers on the discovery of unintended forms of interactions that robots might have with bystanders or passersby. In this view, the naive and untrained responses of people incidentally interacting with the technology are the most important to understand \cite{dix2002beyond}.

For example, \citet{babel2022findings} investigated the use of cleaning robots and their interactions with passersby in German train stations, inspiring designs for predictive autonomous actions in public settings. Lee et al. deploy BubbleBot in public settings to interact with pedestrians, providing guidelines on the robot's motion design in serendipitous interactions \cite{lee2019design, lee2020ludic}. \citet{yang2015experiences} deployed a robotic trash barrel at a campus cafe, exploring socially acceptable robotic behaviors and investigating how individuals interpret the robot's movements. By studying autonomous delivery robots on the streets, \citet{pelikan2024encountering} showed how the robots become a part of everyday street life both physically and socially. 

In \citet{while2021urban}'s comparison of robots deployed in the urban environments of San Francisco, Tokyo, and Dubai, they found strong differences in the rationales and responses to and for robots. The diversity in the rationales for robotic application reflects differences in the economic, social, and political contexts in each of these urban centers, which, fascinatingly, manifest themselves in the attitudes and responses that citizens have to the robots themselves. The 2021 failure of the Sidewalk Labs Smart City "experiment" in Toronto, which famously featured waste-disposal robots running around, highlights the importance of understanding these factors even in the planning of urban technology deployments \cite{Bozikovic_2022, filion2023urban}. The ways that urban robots are taken up---or not---will be related to larger factors that undergird the reasons and context for urban robot deployment, and the way those reasons are communicated.  

\subsection{Sensemaking in Human-Robot Interaction}

What, then, should be the lens that HRI research should take on public robot deployments? This work proposes sensemaking as a lens to apply to these contexts. Sensemaking has been studied in various contexts in the human-robot interaction literature, from robotic coworkers in factories to on-road delivery robots, to investigate how people would adapt to the introduction of robots in different environments \cite{sauppe2015social, weinberg2023sharing}. People have investigated how sensemaking plays an important role in the understanding of social robots by non-expert users, and how this process can improve trust towards robots \cite{papagni2020understandable}. \citet{hoggenmueller2020emotional} unveiled the factors that influence people's understanding of robots' emotional expressions in urban spaces. 
Furthermore, Lyons et al. studied how sensemaking could help restore trust when robots performed unexpected actions \cite{lyons2023explanations}. Sensemaking has also been exploited to study expectations over robots, and how it ultimately leads to decision-making \cite{szafir2021connecting}.

In the book \textit{Sensemaking}, \citet{madsbjerg2019sensemaking} argues that when making sense of a technology, people reason through 'thick' data, relying on context and culture; 
for instance, when robots are introduced in public spaces, the acceptance and perception of robots may vary not only due to robots' appearances and capabilities but also based on users' gender, social status, or cultural background \cite{siino2005robots, lee2014culturally}. Thus, sensemaking is truly situated and embodied and can be the first step for HRI designers to build interaction patterns and blueprints \cite{lee2009designing,kahn2008design}. Sensemaking is hence important for HRI researchers to design understandable and intuitive robots. We feel this is a strong argument for %applying the method of 
focusing on sensemaking in studying public deployments of robots in urban spaces.

\subsection{Resources of Sensemaking}

In interactions between humans, people have been found to draw on a broad range of knowledge resources, which they use to interpret their interaction partners' utterances, to design their own actions, and to make sense of the joint activity \cite{clark1996using}. Specifically, people make use of what is and has been shared with the interaction partner in prior interactions, including the current perceptually available surroundings, but also on what can be considered to be shared communal common ground, like cultural facts, human nature, and 'ineffable background', i.e. factual knowledge that necessarily comes with being in a certain place, for instance. Fischer \cite{fischer2016designing} shows that people, when talking to robots, make use of similar categories as those suggested by Clark  \cite{clark1996using}, attending, for instance to a common language, a common vocabulary, a common perspective, common perception and to a similar cultural background (cf. also \cite{fischer2021same}). Thus, participants attend to certain aspects of the shared situation, even when interacting with robots, and update their common ground with every turn. Recently, Fantasia et al. \cite{fantasia2022making} argue that the "situated and dynamic coordination and negotiation of meanings, intentions expectations and interpretations that human beings experience on a daily basis in their social contexts" is part of the meaningful engagements between humans and robots, yet is not appropriately accounted for in HRI research. 

Similarly, when making sense of new technology products like the Roomba robot, people take more into account than the technology itself -- Forlizzi \cite{forlizzi2007robotic} argues that social dynamics, economics, and environmental issues can play important roles in informing people's engagement with the new technology. From a reversed perspective, Gretzel \& Murphy \cite{gretzel2019making} argue that the anthropomorphism in the individual interactions with a robot takes sensemaking beyond the realm of technology-related ideologies. On the other hand, robots also invite sensemaking that rests on the complexities of discourses around social justice and equity. 

Moreover, in a recent paper on trust in robots, Cameron et al. \cite{cameron2023social} find people to include the 'deployer', i.e. the individual or organization deploying the robot in a given context, in their considerations of the trustworthiness of robots and argue that trust towards robots is happening "{\em within} a human-human interaction social context." They find a substantial role of the 'deployer' in shaping people's attitudes towards, and trust in, the technology.  

To sum up, while previous work on sensemaking in HRI has focused on the iterative, dynamic, and constructive nature of sensemaking in interactions between humans and robots \cite{rudaz2023inanimate}, and while some resources were identified that people draw on when engaging in a sensemaking process, it still remains unclear to date what kinds of resources people who encounter robots in public spaces evoke, how they perceive such robots and how they make sense of their behavior, appearance, and presence.  

\subsection{Previous Work on Trash Barrel Robots}
The deployment of everyday robotic objects in public spaces has been a core theme of our research; multiple research papers have been published on this topic. The first trash barrel robot was deployed about ten years before the deployment discussed in this paper, where a trash barrel was mounted on a Roomba Create base at an on-campus cafe \cite{yang2015experiences}. Through the original deployment, Fischer et al. looked at the initiation and negotiation between the robots and users, highlighting that the ostensive lack of social signals itself is also crucial, a sign of unwillingness to interact \cite{fischer2015initiating}. Ten years later, we brought the project back with upgraded hardware and deployed the robots in open plazas in New York City \cite{bu2023trash}. We published the deployment process, including the ethical approval procedure, and opened a call for similar deployment protocols to promote in-the-wild deployments \cite{bu2024field}. The data collected through the latest deployments are made available upon request \cite{bu2024ssup}. With this dataset, \citet{brown2024trash} investigated the emergent interactions between the plaza users and the robots that naturally come about, where they applied ethnomethodological and conversation analysis to scrutinize the turn-taking nature of the interactions without delving deep into the users' mental models. Different from previous publications on trash barrel robots, this paper focuses on the sensemaking process that the plaza users went through when they came across the robots in everyday settings.

%% file: 03_method.tex
\section{Method}

In alignment with our previous Trash Barrel robot study, we aim to elicit behaviors people have in response to service robots in a public space, without the intervention of signage or design features to explain the robot or how to interact \cite{yang2015experiences,fischer2015initiating}. Based on the prior study, we had some ideas about how people might respond, but since this current study took place in a more public setting and many years after the original study, we did not attempt to make changes in the overall methodological approach. Wizards were instructed to operate the robots around the public space and interact with people passing by and sitting at cafe tables on-site, based on the video from the other study\footnote{Available online at \url{https://vimeo.com/114106601}.} (Details about the instructions to the wizard can be found in Appendix ~\ref{wizard_instructions}). We found that the wizards had to improvise what to do in various scenarios that emerged.

We would like to note that the choice of robots taking the form of trash barrels is a mutual agreement between the researchers and plaza owners through a series of discussions. In fact, the original proposal features chairbots, a fleet of robotic chairs that invite people to sit. However, the moving chairs may be a safety hazard for the general public. On the other hand, the plaza owner points out that a robotic trash barrel would be ideal. People at the plaza usually leave their trash on the table even if there's a trashcan close by (robotic trash barrels have the potential to provide a service); there would be no direct physical contact involved in interacting with trash barrel robots (interactions are safe); and the robots will not replace the janitors onsite, but rather facilitating their work (enabling rather than replacing).

\subsection{Data Collection}

In our Wizard-of-Oz study, two trash barrel robots, one for standard landfill and one for recycling, were remotely controlled by two on-site research assistants \cite{fakeittomakeit}. (This differs from the Fischer et al. study, where only one trash barrel robot was deployed ~\cite{fischer2015initiating}.) 
The robots were instrumented with sensors to record the surroundings and interactions with participants. 

The robots were deployed in late September and early October 2022 over two weeks, as weather permitted. The robots were deployed for one and a half to two hours, starting around 2 pm each day, during the work week. To collect natural human behavioral data, we did not put up any flyers to prime people. Instead, we administered the consent process after the users interacted with the robots.

\subsection{Site}

We picked the southern portion of Astor Place, a popular plaza in New York City, as our study site. The trapezium-shaped area hosts a coffee shop at one corner and has staggered tables and chairs for public seating. The plaza was surrounded by numerous shops, restaurants, and offices, which resulted in heavy foot traffic during the day. With occasional art installations and interactive sculptures, Astor Place was a welcoming place for people of all ages to work, eat, relax, and socialize, which made Astor Place a suitable site to study challenges to autonomous everyday objects in public spaces due to the high variety of use, audience, and physical obstacles.
  
\subsection{System}
The trash barrel robots are visually similar to the robot deployed by \citet{yang2015experiences}, where 32-gallon BRUTE trash barrels are mounted on a mobile chassis. The main difference is that our mobile chassis is powered by recycled hoverboards, which provide more power and higher speed than the iRobot Create-based platform used in Yang's study. This enables the robot to navigate on various terrains in urban cities. 

We delineate robots' roles using the standard municipal color scheme: a blue barrel for recycling and a gray barrel for landfill. We affix vinyl recycling decals to the front and back of the blue barrel to emphasize its role. The mobile platform is propelled by a re-engineered hoverboard attached to its center. The original hoverboard PCB is switched out for an ODrive v3.6 motor driver controlled by a Raspberry Pi 4 (RPi4) computer.

\subsection{Data}

A 360\degree ~camera was mounted in front of each barrel above the edge. People are aware of these cameras at first glance. The camera records both 360\degree ~video and audio of the activities all around the robot. Two standard GoPro cameras were mounted high up on the wall of the coffee shop at the plaza's southeast corner, each covering half of the plaza, mostly unnoticed by the participants.

\subsection{Wizard}
For each deployment, two members of the lab controlled the robots as the hidden wizards on-site. Each wizard controlled one robot through a remote joystick for non-verbal behavioral processing. The wizards were trained in the lab before deployment to familiarize themselves with the joystick controller and practice how to navigate the robots in crowded spaces. During the study, wizards sat near the edge of the public sitting area and were encouraged to communicate with each other to keep them aware of the robots' surroundings. Wizards were instructed to keep their robots socially appropriate. For example, the robots should not run into tables or passersby, should not accelerate towards people at short distances, and should respond to people's requests accordingly. The robots could also "offer" their service if the wizards thought it appropriate (e.g., a person is taking the last bite of their burger). In addition, the wizards were instructed to keep the two robots close to each other during the deployment.

\subsection{Interviews}
Besides the two hidden wizards, another on-site researcher was in charge of interviewing people after they interacted with the robots.  The researcher would approach people and invite them to participate in a very short semi-structured interview regarding their opinions towards the robots. The participation is completely voluntary. The on-site researcher tried their best to interview as many users as possible, but it was unrealistic to interview everyone, given that interactions were happening frequently.  
The interview questions were designed to be maximally open to allow participants to provide their most honest feelings towards the robots. 
Thus, we asked them what words came to their minds when they thought of the robots, what things they noticed that they approved or disapproved of, how they would rate the robots, and whether there was anything else they wanted to say -- see Appendix \ref{questions} for the interview script, and Appendix \ref{consent} for details on the relevant permissions and consent procedure.

%% file: 04_analysis.tex
\section{Analysis}

To understand people's sensemaking activities when encountering the two trash barrel robots, we pursued two paths to data analysis. First, we performed a thematic analysis on the transcripts from interviews of people who had interacted with the trash barrel robots, to see what topics people raised when we asked them about the robot. See Appendix \ref{questions} for the interview script. Second, we performed a video analysis to find moments of observable sensemaking occurring between members of the public and the robot. 
Methodologically, we chose to separately analyze the interview transcripts and interaction footage because people's answers from interviews may not always strongly correlate with their behavior. In addition, due to the nature of the data collection process, we have more video footage than interview data. However, while we begin with analyzing the interview and video footage independently, we try to link data sources at a per-interaction level, if possible, to better ground our findings. 
This mixed-method approach helps us to identify both overt sensemaking behaviors occurring in the public setting and also to discern less obvious sensemaking that people engage in as they watch and reflect on the robots in their space.

\subsection{Interview Analysis}
\subsubsection{Protocol}
To analyze the interview data, we use a qualitative thematic analysis approach called 'affinity diagramming' \cite{lucero2015using}. In this bottom-up approach, first, all interviews are transcribed and printed, the interview responses are identified and excerpted, and then the responses are organized in several rounds of coding to develop themes. Specifically, all responses were cut apart so that each response to an interview question was taken as at least one data point; if participants mentioned different topics, they were treated separately. Next, all the segmented interview responses were assembled based on the similarity of topics in a bottom-up iterative process. After clear categories had emerged, all data were coded by two different researchers using these categories. That is, we subsequently used the themes to code each of the responses, such that some responses could be included under multiple themes. Two researchers separately coded the transcripts based on the developed themes. Since the themes are not mutually exclusive, we compute Cohen's Kappa coefficient for each theme individually and average the coefficient across themes. The resulting multi-label kappa coefficient is 0.62, which shows substantial similarity \cite{hallgren2012computing}. 

\subsubsection{Themes}
Here we describe each theme, ordered by frequency, and illustrate exemplar quotes in each. Some responses spanned multiple themes. For example, the following quote touches on five themes: the robots, the concept, people, trash, %\anon{New York}
the city, and participant sentiment: \textit{"Well, it was startling [sentiment] to see nobody with them. And then I thought, what a good idea. [concept] So people can be putting their trash in because New Yorkers are [the city], sometimes they leave trash at the table. [trash] So I thought too, 'Yeah, it's really cute.' [robot]" }

\paragraph{Robot Design} In this theme are interview statements that focused on the robot's design (\textit{"It just looks like a regular trash can with the camera on it."}, or adjectives to describe it (\textit{"wonderful," "creepy," "cute", "dystopian," "awkward," "adorable"}). Some of the comments were somewhat anthropomorphic, using terms that are usually applied to humans, like \textit{friendly, nice} and \textit{beautiful}. Multiple people suggested that we accentuate the anthropomorphism: \textit{"I feel like you should add googly eyes."} Other comments were more in line with the way we talk about objects, focusing on affordances or the experiences they evoke, such as {\em fun} or {\em amusement}, {\em cool}, {\em fantastic}, or {\em awesome}. 

\paragraph{Role/Idea} The role theme pertained to the function, job, or purpose the robot was fulfilling: 
\textit{"I thought it was going to pick up trash, but then it didn't. It was just walking around." }
Sometimes people evoked other robots--Wall-E, R2D2, Roomba--in these statements. There was some variance in how people interpreted the robot:  \textit{"Obviously, it's a trashcan, so put the trash in."}
\textit{"It's like I don't think of them as trash cans even." }

Interestingly, many interviewees gave feedback not on the robot itself, but on the \textit{idea} of the robot: \textit{"I think it's, um, I don't think it's necessary." }
Examples are \textit{"The concept is good, very good in fact: It's modern, it's ecological and everything."}; \textit{"I like the idea,"}; \textit{"It's very nice and I think it's recommendable."} and \textit{"It's a reminder. It's a reminder that we're all working together to keep the place clean." }

\paragraph{Interaction with the Robots} The interaction theme focused on the robots' behavior and on the exchanges the robot had with participants.
We included both descriptions of actual interactions, and speculations on interactions offered by the participants: \textit{"You signal it, right? It comes to you. If you're lazy, you'll still be able to throw out the trash."} \textit{"It was active. It wasn't just standing, still moving around. It was perky."} \textit{"Um, um, you should make it say swear words sometimes."} \textit{"It's a little weird, so I wouldn't be surprised if somebody just felt like kicking it."}

Early on, people seemed not to know if the interaction was intentional or accidental:
\textit{"...I saw it in the corner of my eye. I thought maybe I was tired, and then I was seeing things."};

Another person commented, \textit{"At first you see it moving, because I thought it was just the breeze blowing, and I'm like, wait, no, it's moving. At first, you're like, okay, it's moving, it's a robot."}

\begin{figure}[htbp]
    \centering
    \begin{subfigure}[t]{0.68\textwidth}
        \centering
        \includegraphics[width=\textwidth]{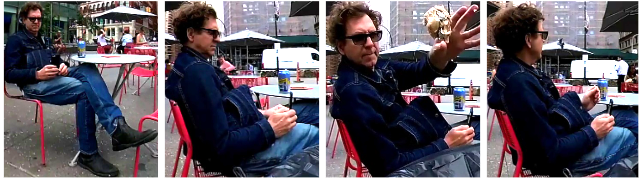}
        \caption{$P_{prag}$ assumed robots acted for themselves.}
        \label{fig:prag}
    \end{subfigure}%
    \hfill
    \begin{subfigure}[t]{0.29\textwidth}
        \centering
        \includegraphics[width=\textwidth]{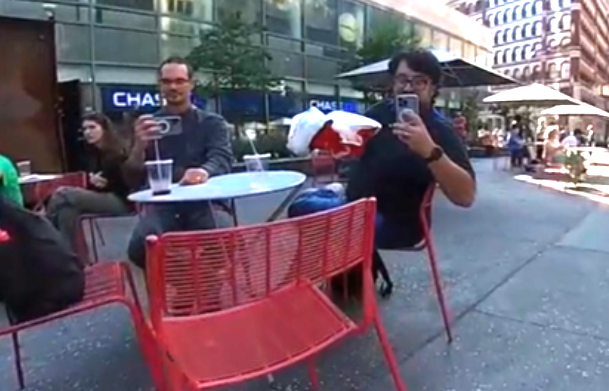}
        \caption{$P_{impe}$ waved trash at the robot.}
        \label{fig:impe}
    \end{subfigure}
    \caption{People's assumption of whether the robot is pragmatic or imperative may influence their interaction strategy with the robot.}
    \label{fig:overall}
\end{figure}

Once people confirmed that the trash barrels' movements were intentional,
it seemed they either assumed the robots acted for themselves (pragmatic) or demanded people's responses (imperative).  
People who assumed the robots were pragmatic stated, \textit{"I like that they were kind of making their way around and seeing who had what and stopping by tables."}; \textit{"Well, I mean if they were running around all the time, it would be a little disconcerting" ($P_{prag}$).} On the other hand, people who thought the robots were imperative commented, \textit{"and it moved on... I knew exactly what I was supposed to do if I [had] trash to put it in."}; \textit{"They drive to people and ask for trash}"; \textit{"I didn't know if they [could see] me or acknowledged me or if I should just go and do something else" ($P_{impe}$)}. 
These assumptions can be reflected in the ways people interact with the robots. When $P_{prag}$ interacted with the robot, he sat in his chair looking down at his phone. He did not show any sign of interaction until the robot drove past him, quickly tossed trash into the robot, and resumed his normal activity right after (see Fig. \ref{fig:prag}). He assumed the robot had its own plans and, hence, did not need to actively engage with the robot. On the other hand, $P_{impe}$ actively waved trash at the robot, trying to get its attention (see Fig. \ref{fig:impe}).

\paragraph{People/Society} The interviewees often discussed \textit{"people," "somebody," "you,"} or \textit{"nobody"} in the abstract when they discussed both the interactions, but also the need for trash pick-up, or changes in society: \textit{"I think the trash can is a good idea because it makes people think."} \textit{"It's brilliant, but as we say, you know, nobody's going to have jobs."} \textit{"I mean, how much lazier can you get?"} \textit{"They can spend money more for people than the trash".} There were also concerns about homeless people, people who have autism, and people in wheelchairs -- both ways, such that the robots could assist them but also hinder them. 

\paragraph{Purpose of the Deployment} The deployment theme focused on what the robots were doing in the public space. Since the participants did not know \textit{why} there was a robot in their environment, many of the statements focused on speculation: \textit{"I don't know why it's like that. Before you came, we were asking, is it a joke?"} \textit{"Is that an idea that they have going forward to maybe have remote-controlled garbage cans? Or just an idea to see how people feel?"} \textit{"My first thought was, oh, it's some kind of art project."} \textit{"I guess I'm wondering, does the data get, like, the footage get saved anywhere? Is it like
rogue going around?"}

There were also many questions directed at the interviewer: \textit{"Is it just for school? Awesome."}; \textit{"That was you who created that?"}; \textit{"Yeah, what's the organization?"}

\begin{figure}[h]
  \begin{minipage}[c]{0.4\textwidth}
    \raggedright
    \includegraphics[width=\textwidth]{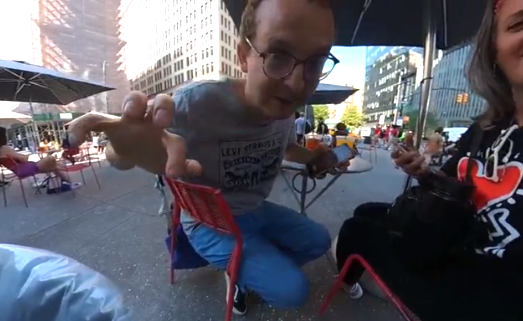}
  \end{minipage}\hfill
  \begin{minipage}[c]{0.55\textwidth}
    \raggedleft
    \caption{
       \textit{"I don't know why it's like that. Before you came, we were asking, is it a joke?"} While not always obvious, people's interactions reflect their beliefs about the deployment of robot trash barrels. Here, a French couple who were unfamiliar with the signage in the U.S. thought the deployment was a prank, and they played along. \textit{"Why one is blue, one is grey, for which one is it, you know, recycled or not recycled. We don't know. Me personally, I don't know. I don't know the difference between the two of them."}
    } \label{joke}
  \end{minipage}
\end{figure}
\paragraph{Sentiment} Sentiment included statements about how participants said they felt about the robot, its interactions or the deployment, what they said they were thinking, or what they speculated other people felt: \textit{"If it was like smooth rolling the entire time, I would be very creeped out."} \textit{"I never wanted anything more than a trash can."} \textit{"Do people get scared? Do they get up and move?"} Sometimes the sentiment was implied: \textit{"F*cking killer. Thank you."}

At the same time, seven people described their experience as "interesting," and five people indicated some kind of shock, e.g., \textit{"speechless"} or \textit{"just shocked."}
Furthermore, nine people used terms like \textit{"kind of freaky,"} \textit{"a little weird,"} \textit{"strange,"} or even \textit{"scary."} 

\paragraph{Trash}
People talked about the trash that the trash barrel collected, but as a general problem rather than in specificity:  
\textit{"It's pushing people more to recycle or throw trash out. Especially in New York. It's a big city.}" 
\textit{"I thought it was going to pick up trash, but then it didn't. It was just walking around."}  \textit{"I like this idea too because it prevents littering."}  \textit{"I like that there were two of them. One was for trash and one was for recycling."}

\paragraph{Cameras and surveillance}

Fourteen people raised privacy concerns given the cameras mounted on the robots. For instance:%\\ 
\textit{"I'm wondering, does the data get, like the footage, get saved anywhere?"}; %\\
\textit{"It's kind of intrusive and perhaps non-consensual with like the camera aspect of it,"}; %\\
\textit{"It's a little surveillance-y."  }

\paragraph{Other topics}
Additional topics we coded that had fewer counts were:
Other Interaction (how the robots interacted with the physical space or each other), [The City], Robot Hardware/Software/ Technology, and Money.

\subsubsection{Results}

\begin{figure}[h]
  \begin{minipage}[c]{0.7\textwidth}
    \includegraphics[width=\textwidth]{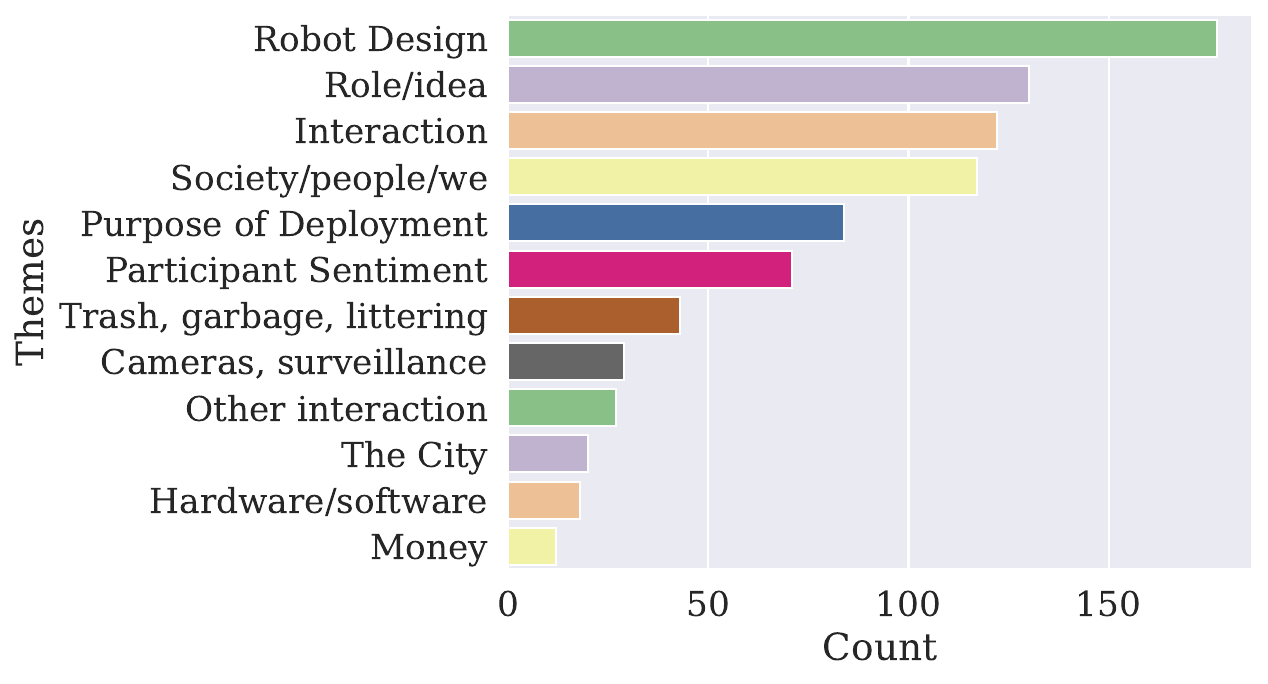}
  \end{minipage}\hfill
  \begin{minipage}[c]{0.3\textwidth}
    \caption{
       Results from thematic coding on the interview transcriptions. The x-axis represents the number of sentences from the interview that are coded with the corresponding theme.
    } \label{fig:03-03}
  \end{minipage}
\end{figure}

While the absolute counts in each category have little importance, the relative weight of each theme shows what aspects of the robot encounter are striking to them. They notice and comment on the robot itself, but also, to a large degree, communicate about the robots such that they are taken to stand for the underlying idea or concept that the robots embody. They focus on the mechanics of the interaction and on how to beckon or shoo away the robot, but they are also curious about the deployment and why the robot is in their space. 

Many people (\textit{N=34}) report on processes of sensemaking directly. For instance, they report on initial confusion, for example when they first see a trash can move, and how they overcome this initial confusion, for instance:\textit{"I didn't know if it was just an experiment or if I'm allowed to put something on there,"} or 
\textit{"...and then you go like, What? and you want to use it,"} or 
\textit{"Once you kind of get used to it, it's fun just waving it over."} or
\textit{"We didn't know whether to speak to it or wave it over or what."} 

Several people state that they were uncertain about how to interact with the trashcans, e.g.: \textit{"There's a little more information on how I'm supposed to interact,"} \textit{"If there was more of an indication as to what the purpose is,"} or \textit{"We don't know how to do that. We don't know how to coax a garbage can to me."} Many other people report on their hypotheses about how the robots work, for instance, or\textit{ "It was funny that it followed me like a dog,"} or \textit{"It responds really well to my hand signals,"} or \textit{"It was coming when people beckoned it and were like, hey, I have trash,"} or \textit{"I like how it kind of felt like it was interacting with us,"} or \textit{"It seemed like the can cared about me doing what I wanted to."}

Finally, people make many suggestions for improvement of the robot design and behavior, asking to add dialog capabilities or anthropomorphic features (e.g. \textit{"maybe with eyes"}), and they start brainstorming about (other) applications of the robots, like helping the homeless, blind, or autistic people. Furthermore, there are comments that concern the commercialization of the robots, like \textit{"You could sell it to the city"} or \textit{"make a lot of money?"} or \textit{"it could be positive for someone like you."}

To sum up, there are comments that concern some of the sensemaking processes that have been heavily discussed in previous HRI work, where people draw on animate beings to make sense of robots' behavior; these are comments that evaluate the robots using adjectives or attributions from the human domain, like {\em cute}, {\em adorable} or {\em friendly}, or when talking about {\em feeding} the robot etc. However, the majority of the interview responses do not concern one-on-one interactions with the robots, even though there are many comments that explicitly refer to active sensemaking activities. In many cases, people consider the robots {\em an idea} or {\em concept}, and they spend much effort on understanding how they are operated, who the people behind the deployment of the robots are, who is going to see the video footage and what role the city may play in this. In addition, they make use of cultural common ground when making sense of the robots, like when referring to other robots or to cultural knowledge concerning the recycling symbols. Thus, people see far beyond the one-on-one human-robot interaction and attempt to make sense of the robot being deployed in a certain place by certain people for certain social purposes. 

\subsection{Video Analysis}

\subsubsection{Protocol} We analyzed video footage captured from both on-robot and in-situ cameras for instances of sense-making activity. Two researchers, one was the interviewer and the other a wizard during the deployment, went through the videos and annotated every interaction, both explicit and implicit, between the robots and people at the plaza. For each interaction, the researchers logged the time, which robots were involved, the group size, and a brief description. Two other researchers, who were not involved during the robot deployment, went through the annotations to refine the timing and description further. This is to make sure the annotations are clear to the general public. 

The robots were out in the field for 10 hours, producing a total of 40 hours of footage from all cameras. We focused on the 20 hours of footage captured by the onboard cameras, where interactions were dense and captured from a first-person view. In total, we identified 274 interactions, from which we made 166 sub-clips for detailed analysis. 

To focus on the sensemaking activities, the research team scrutinized the videos for instances of observable sensemaking processes. Subsequently, the team re-viewed the clips to develop distinctions between different types of sensemaking activities. 

\subsubsection{Vignettes}
We picked out three video clips to showcase how sensemaking was performed during the deployment, i.e., moments where people have extended engagement with the robot and with other bystanders and observably engage in sensemaking of the robots' behavior and function. We narrate each of the encounters below, and the original video footage is included in the supplementary material. 

This qualitative selection of sensemaking behaviors is not an exhaustive list but is intended to provide fine-grained illustrations of the ways that sensemaking activity unfolds.  

\newpage

\noindent\textbf{Clip 1: Experimentation}

\noindent\begin{minipage}[h]{0.65\textwidth}\raggedleft
\begin{verbatim}
Two girls are sitting at a table facing each other. 
[Girl A sits on the outside, and Girl B sits across from her.]
Girl A starts to film the robot, while Girl B starts waving 
to call the robot over. 
As the robot approaches them, Girl B reaches for her 
backpack, looking for things to throw away. 
Girl B passes Girl A some trash and says, "Throw this 
sh*t out."
Girl A ignores her and says "Hello" to the robot 
while waving at the robot.
Girl B repeats herself, but Girl A keeps waving. 
Girl B says, "It's not going to wave back to you, it's not." 
In the meantime, Girl A finally takes the trash 
from Girl B.
Girl A turns to the robot and asks, "Can I throw this in?"
while raising her hand to show the trash.
Girl B says, "Yeah."
Girl A shows the trash to the robot. 
Girl A peeks into the trash barrel and raises her hand to 
show the trash to the robot again. Then she retracts 
her arm.
Girl B says, "Are you scared of a moving trashcan?"
Girl A chuckles and says, "No."
Girl A tosses the trash into the barrel.
The robot prepares to leave and shakes a little bit.
Girl A looks at the robot and asks, "No?"
As the robot backs up and drives away, both girls say, 
"Thank you."
\end{verbatim}
\end{minipage}%
% \hfill%
\begin{minipage}[h]{0.35\textwidth}
\raggedleft
\includegraphics[width=0.8\linewidth]{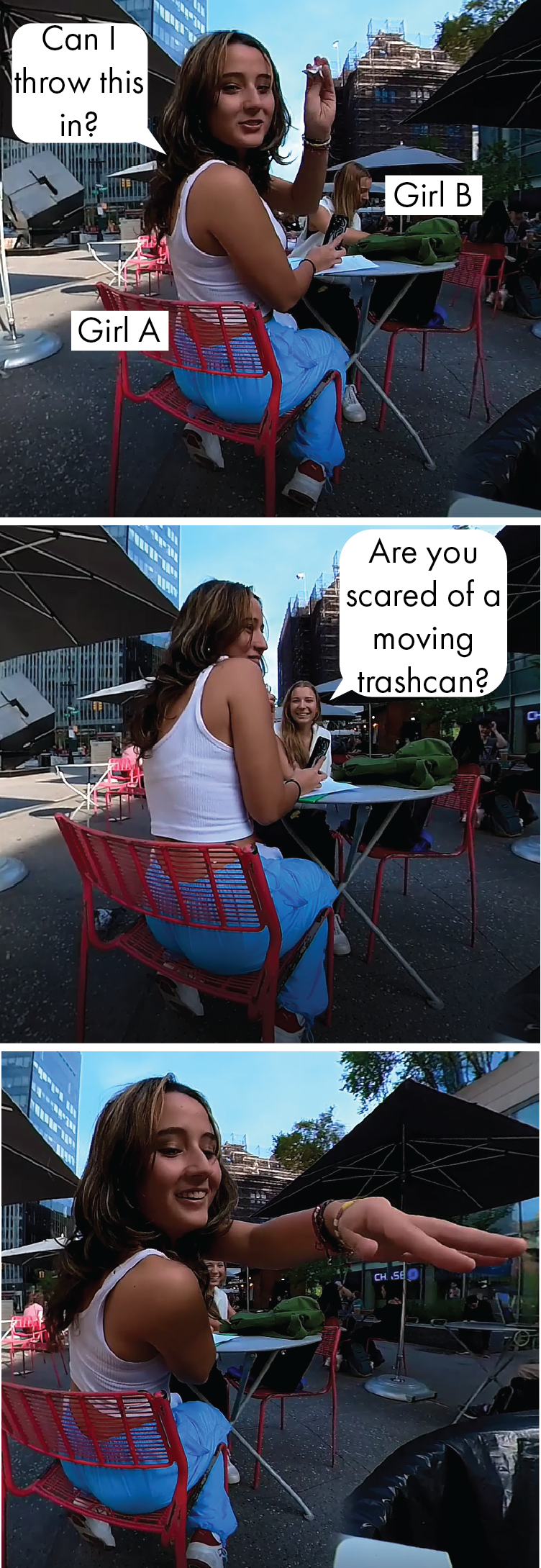}
\end{minipage}

\vspace{1em}
In Clip 1, we can observe sensemaking happen during the interaction between Girl A and the robot. The requesting action performed by Girl B is treated as natural by all parties at play. While the robot approached the table, Girl A was still waving at the robot and expecting more feedback until Girl B interrupted her, saying the robot was not going to wave back. Girl A kept trying to interact with the robot by asking a question about whether she was allowed to throw her trash into the robot. Her subsequent behaviors---peeking, raising her trash, and then retracting---showed that she was waiting for confirmation from the robot, displaying uncertainty about the functionality of the robot. It was not until her friend made fun of her that she finally threw out the trash. The peeking action showed that she was looking for reassurance from other trash in the barrel that the robot was usable. But once she threw out her trash, she misunderstood the delay in the robot's response as a sign of trouble. The girls ended their part of the interaction by saying "thank you" only after the robot made a clear leaving gesture. Throughout the interaction, Girl A monitored the robot's subtle motions, visibly trying to make sense of the meaning behind them.

In the subsequent interviews, both girls verbally described their hesitation, confirming what we observed from the video vignettes. \textit{"I didn't think it would, like, recognize it. Yeah, I didn't think it would want to come. I thought I was like, I could try. I don't think it's going to come."}

\newpage
% demonstration.mp4
\noindent\textbf{Clip 2: Demonstration} 

\noindent\begin{minipage}[h]{0.65\textwidth}\raggedleft
\begin{verbatim}
[The landfill robot gets stuck and struggles to get out.]
A girl walks towards the robot and stops when she 
realizes that the robot is moving. 
She then approaches a man standing next to the robot and 
asks, "Whose trashcans are those?"
The man replies, "I have no idea, bro."
The girl walks a few steps forward and stops to 
observe the robots again. 
[The recycling robot bumps the landfill robot to get 
it unstuck.]
The girl says, "Are they fighting? Hello."
A woman approaches from the back of the robot with trash 
in her hand.
[The woman previously had interacted with the robots.]
The woman says, "That's a trashcan," and then she looks 
at the girl briefly.
Then, the woman turns to the robot and says, "I have some
trash here. Here. Can I put it in?"
[The landfill robot approaches the woman.]
The woman says, "Is that a yes?"
[The landfill robot approaches the woman more.]
The woman says, "Haha, it's a yes." 
She then put her trash into the trash barrel. 
The woman says, "Thank you! Very nice service you are 
doing. Thank you very much. Now stop hitting your friend."
\end{verbatim}
\end{minipage}%
% \hfill%
\begin{minipage}[h]{0.35\textwidth}
\raggedleft
\includegraphics[width=0.95\linewidth]{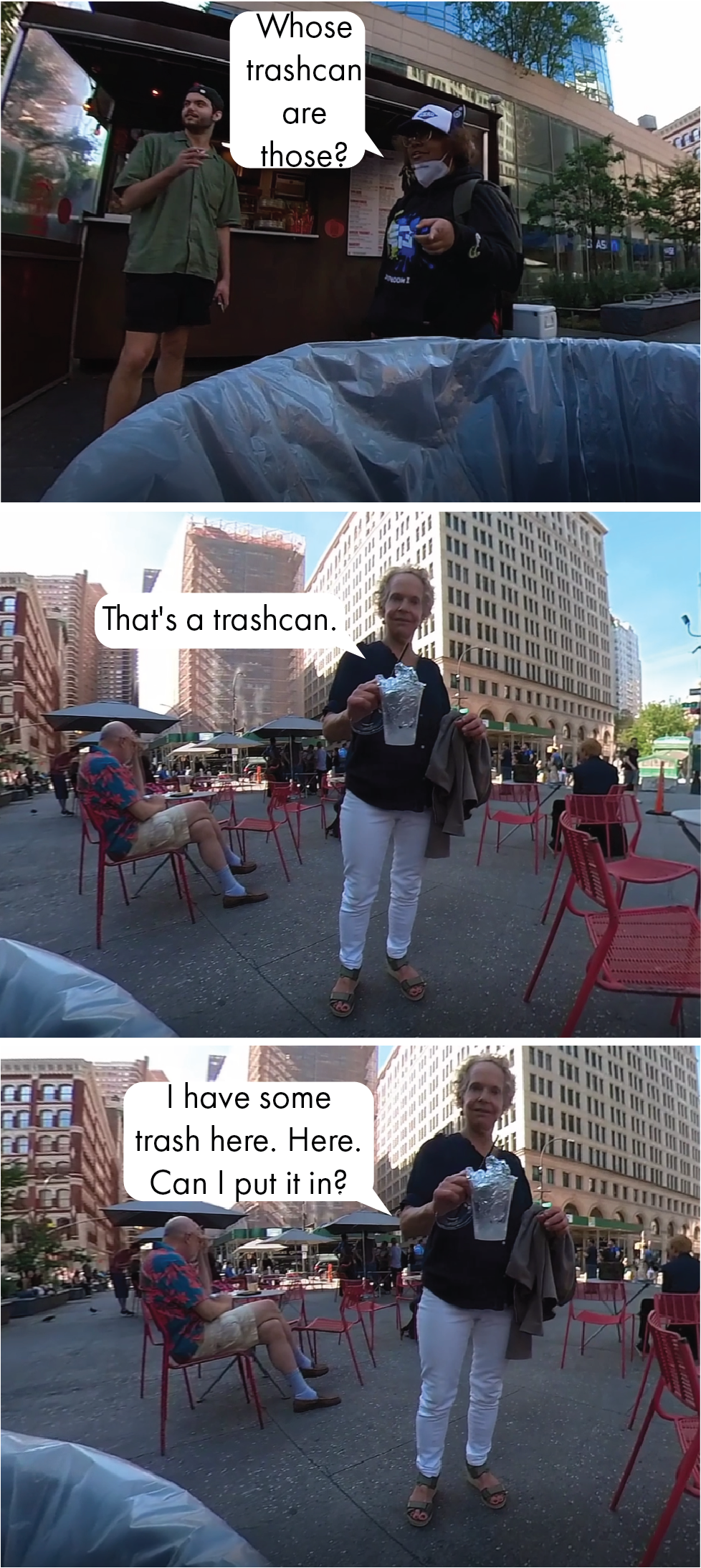}
\end{minipage}

\vspace{1em}
In Clip 2, the girl first approached other bystanders to inquire about the ownership of the robots, which is consistent with the \textit{deployment theme} in the interview analysis. Then, the girl kept observing the robots' behavior and tried to decipher the meaning of their motions. However, she misunderstood the "bump to rescue" behavior as robots attacking each other. When the woman approached later, the woman was clearly demonstrating the usage of the robots to the girl. She first pointed out that the robots were still trash cans. Then, she elaborately had a "conversation" with the robot to show that the robot has intelligence. Throughout the interaction, the woman and the girl never spoke to each other directly, but the usage of the robots was demonstrated and shared. It is worth mentioning that the woman had interacted with the robot before this encounter, where she also had a brief conversation with the robot (she said, "Want some trash? There, there's some trash.") Note that she does not wait for the robot's confirmation as she did in this encounter. To demonstrate the robot and help the girl make sense of the robot, she modeled and narrated every single step of the interaction she felt was proper, and looked to the robot for confirmation. 

A close analysis of the woman's interview transcripts revealed indications of her demonstrative behavior. \textit{"I thought what a good idea. So people can be putting their trash in because New Yorkers are, sometimes, they leave trash at the table... It's really cute."} The lady supports her interpreted idea behind the deployments, which gives her incentive to promote the concept.  She added, \textit{"I knew exactly what I was supposed to do if I [had] trash [to] put it in"}, reflecting her belief that her actions were appropriate.  Her understanding of the robots' affordances and belief in the concept provided her with both the rationale and the confidence to demonstrate the robots' use to others.

\vspace{1em}
\noindent\textbf{Clip 3: Antagonism}

\noindent\begin{minipage}[h]{0.45\textwidth}\raggedleft
\begin{verbatim}
A man sees the landfill robot and says, 
"What the f**k?" 
He then follows the robot.
Whenever some pedestrians 
walk past him and the robots, 
he will point at the robot and 
speak to the pedestrians.
The man points at the robot and asks, 
"Who the f**k is  doing this?"
The man is standing in place, constantly 
looking around.
After one pedestrian says, "I love them," 
he walks toward the robot and 
pulls the robot by the side 
to make it fall to the ground. 
He says, “Not that sh*t.”
He then walks away but, after a few steps, 
stops to look back. 
\end{verbatim}
\end{minipage}%
\hfill%
\begin{minipage}[h]{0.5\textwidth}
\raggedleft
\includegraphics[width=0.9\linewidth]{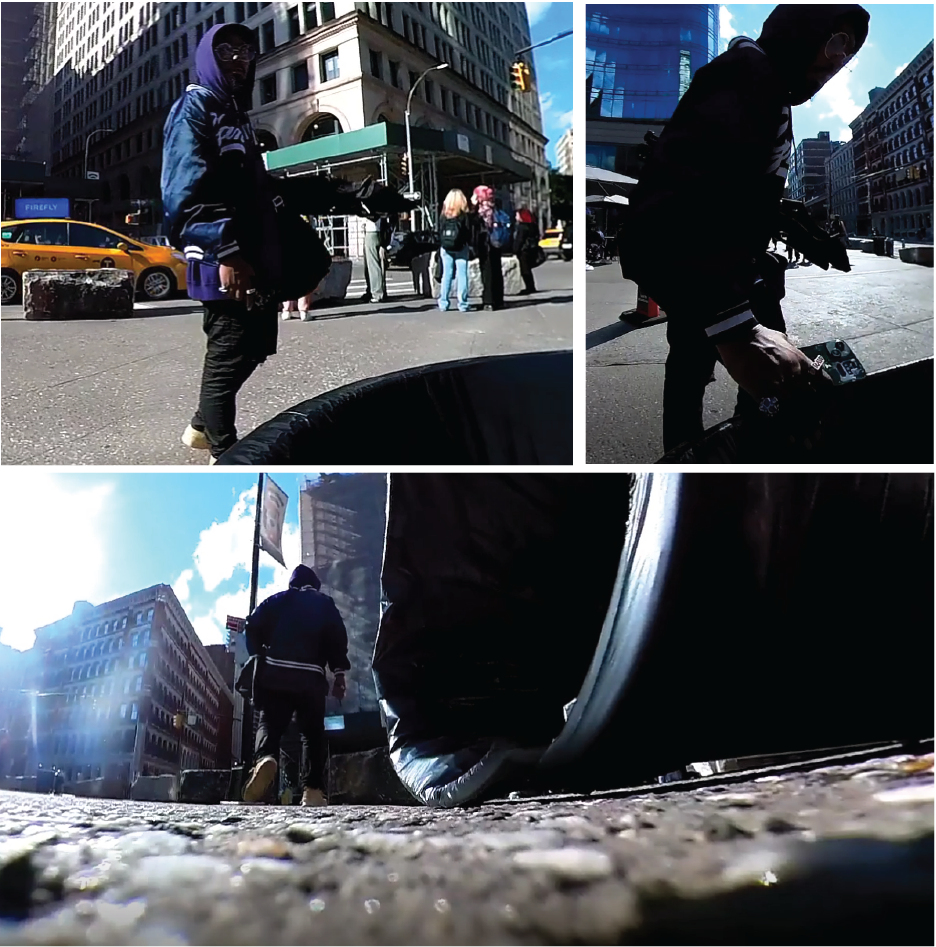}
\end{minipage}
\vspace{1em}

Clip 3 demonstrates a violent encounter with the robots in a public place. It is clear that the man was trying to figure out the ownership of the robot. He asked all the pedestrians he encountered about the trash can. While his voice was mostly obfuscated by the robot's motor sound, it is reasonable to assume that all his inquiries were about the ownership of the robot. From his aggressive behavior and language, we assume that he associated the robots with public surveillance, which is a common concern among participants.

\subsubsection{Results} 
Unlike the interview analysis, which largely shows what people are trying to make sense of and features a diverse range of themes, the video analysis captures how people perform sensemaking in action. 

The video provides evidence of confusion that we also see in the \textit{interaction} theme. As shown in Clip 1, when new users interact with the robots for the first time, their action tends to be exploratory and hesitant. We observed many instances of coaxing action, such as waving, holding trash, and whistling, where people tested out the robots' affordances. Some people intentionally blocked the camera to see how the robot would respond to it, others tried to hold a conversation with the robots directly by inquiring if the robots wanted trash, despite the robots showing no capability of conversing. Once people initiated the interaction, they naturally assigned meaning to the robots' subsequent motion, even though sometimes the hidden wizard did not recognize their requests. For example, people would interpret the robots' wobbling motion as a 'yes' to their question, even if the wobbling was merely caused by the bumpy surface. The girl in Clip 1 perceived the robot's slow response as a rejection of her trash, even though the delay was rather caused by technical issues. Once the participants familiarized themselves with the robots, they were willing to demonstrate or even perform the usage of the robots to other people. As shown in Clip 2, the woman had interacted with the robots before, and she deliberately demonstrated the interaction for the girl. 

The other theme that can be supported by visual evidence is \textit{deployment}. To some pedestrians, when they first encountered the trash barrel robots, their first reaction was to look around to find out who the robots belonged to. As shown in Clip 3, when the man spotted the robots, he immediately looked around and asked passers-by who was in charge of the robots. The \textit{deployment} of robots was also a common topic among serendipitous conversations triggered by the robots' presence, where strangers gathered together to make sense of the purpose and owner of the robots. 

\textit{Sentiment} was also easily recognizable from video clips, where people who liked the robots were observed to hug and play with the robots, while people who were not fond of the robots would verbally or physically express disagreement. However, different from the interview analysis, the reasons for their sentiments could not be observed from the footage. 

In summary, the video analysis provides visual evidence for some of the themes that emerged from the interview analysis. 
The footage shows people exploring, poking, probing, testing, and openly discussing the robots and their deployment in this specific place.
For other themes, the video footage mostly captures the result of sensemaking processes, rather than their source. For instance, when a person decides to welcome the robot with a hug or kick it with their foot, they have already gone through the sensemaking process of why the robots were deployed here and their roles. Video analysis could not then capture their sensemaking process and what resources they based their reasoning upon.

%% file: 05_discussion.tex
\section{Discussion}

Our analysis of the interviews and the videos has shown that to make sense of the robots, people take a broad range of context factors into account. Among those are the deployers, i.e., the people behind the robot deployment in the specific space (cf. also \cite{cameron2023social}). Concretely, in their evaluation of the robots and their experience with the robots, the participants' understanding of who is operating it, who put it there, who profits from it, and who gets to see the video footage plays a crucial role. The insecurities arising need to be taken into account -- preferably in the design of the robots, since the robots are probably encountered more readily than other possible signage in the environment. That is, designs of robots for the field should provide implicit answers to people's questions that arise.

Beyond the concrete deployment, people were found to draw on a large range of resources, from a shared cultural background, which included (common realizations of) city policies, local recycling practices, and culture-specific symbols, via human nature, to personal common ground from experience with the robots themselves, in line with the categories of common ground described by Clark \cite{clark1996using}. With this, our findings go beyond the findings by Cameron et al. \cite{cameron2023social}, who raised the role that the deployer of a robot has on the trust within human-robot interactions; our work shows that it is not the deployer as a stand-alone entity that needs to be considered, but a far wider cultural and contextual context the interaction takes place in, of which the question of "who is deploying the robot" is just one question.

At the same time, our results show that many of the processes identified to be relevant in one-on-one human-robot interactions are only relevant to some extent in our field scenario; for instance, while many were very excited about the robots and talked about them in emotional, anthropomorphizing terms (e.g. {\em cute} or {\em sentient}), many more considered them as a good {\em idea} or {\em concept}, thus not focusing on the robots as such, but as instantiations of a concept or "reminders" of a particular practice. Several also just referred to them as trashcans that happen to move, and were discussing the value of trashcans moving around compared to static ones. In line with this view is also the lengthy discussion about how the robot deployment may make people lazy or destroy jobs -- which in turn is in line with Gretzel \& Murphy's \cite{gretzel2019making} suggestion that robots, in comparison with other technologies, invite considerations about social justice and equity. 

\subsection{Public Sensemaking}

\subsubsection{Sources of Sensemaking}
To make sense of the robots, people in public spaces draw on different resources. They look for information beyond the space, and they ask questions aloud of one another; they also experiment with the robot, talking and asking it questions even though it does not respond. We also see people watching one another interact with the robot and commenting on each other's interactions. When they feel they understand the interactions, they take it upon themselves to perform demonstrations to help other people know how to interact.

 \subsubsection{Sensemaking of Behavior}
From the moment that they glimpsed the robots through the corner of their eyes to the moment they threw trash into the robots' barrels, people went through the sensemaking process to form an understanding of the robots and piece together the affordances of the robots. 

 \subsubsection{Sensemaking of Deployment}

While the behavior and interaction of the participants in the videos show how they are making sense of how to interact with the trash barrel robot, the interview data helps us to understand some of the questions they are trying to ask beyond the interaction. They are trying to map the specific instance of the robot they are seeing to the larger role or concept of robots that they have heard about. They are also trying to understand the implications that the robot they are seeing has on society. For example, they think about the implications for cleanliness and public health, of impacts on workers or vulnerable populations, or the implications of the surveillance and data collected by the robots. 

Sensemaking of deployment usually happens before sensemaking of behavior, because the framing of the interaction is integral to the way they interpret it. As we observed in the video analysis, people's interaction behavior varies drastically if they perceive it as a commercial prototype, as a part of public infrastructure, or as an art project. Without a clear understanding or assumption of the circumstances of the robots' deployment, people may be hesitant to act. Thus, sensemaking of deployment is a precondition of sensemaking of behavior and is often ignored during the design process. 
 
\subsection{Implications for Design}

What this study makes clear is that, when designing robots for public deployment, it is important to think about all of the other questions and issues that people encountering the environment will be trying to make sense of. This goes beyond the form, interface, controls, movements, behaviors, and interaction of the robot, which have been the traditional remit of HRI research and development. People in public spaces want to understand who is deploying the robots, and why, and that understanding changes whether and how they want to interact with the robot. In this section, we discuss some of the implications for design emerging from our analysis of the videos and interviews from our deployment.

 \subsubsection{Design for Sensemaking}

Because this robot deployment was a study to understand how people make sense of robots in the absence of information, one of the most obvious things to do is to add the information that was missing. What this research exercise helps us do, however, is to understand specifically what kinds of information are needed to support the sensemaking processes that people will engage in. Signage, iconography, or messaging that helps to indicate what the trash barrel is for, do's and don'ts for engaging the robot, who the robot belongs to, who services the robot, and where else these robots are being deployed would have helped answer some of the questions that people encountering the robots had.

By and large, people did not seem to have technical questions about the trash barrel robot: we did not find evidence that people were asking how the robot worked, or how fast the robot could move. A few people asked how long the batteries last; more asked how long the trash barrel would be in the square. They also did not ask questions about what would happen to the garbage or recycling--how often the barrels were emptied, who would empty them, and whether or how the recycling would be sorted. People did, however, have a lot of questions about the 360\degree ~camera on the robot--if the robot used that to navigate, whether it was recording, and who else could see the video. The complexity of possible responses to the question of what happens to the data leads us to feel that there should be icons to signal how footage is processed after it is captured, much as the recycling symbol indicates to people what happens to waste products after they are deposited in each can.

\subsubsection{Design of Robot Behavior}

Designers of the robot have some decisions to make about what they want to communicate about the robot through its behavior. For instance, a robot that is meant to be more incidental in the environment does not need to engage participants at all--it can circle a space automatically and merely draw attention to the affordance of trash disposal, or make the task of moving trash to the trash barrel a little easier. On the other hand, a robot that is actually intended to actively solicit trash would need to find potential users and approach them. A more imperative trash barrel might have fewer, more pointed movements, to assert social power and insinuate "orders" to visitors in the square. 

Pragmatically, the trash barrel robot has no mechanisms of enforcement, no way to "make" people use the trash barrel or clean up after themselves, but the sensemaking perspective helps illustrate that indicating the intended purpose can help people know how to "fall in line" if they want to.

\subsubsection{Design of other contextual elements}
Beyond designing the looks or the actions of the robot, it is also important to consider other elements in the environment that can aid in sensemaking. Because the interaction unfolds over time, people often learn about how to interact with the robot from watching one another. From that perspective, then, building in exemplar interaction moments--from a confederate, or from the cafe stand owners or janitors in the vicinity--can help people to make sense of the new technology and how to engage or not with it.

\subsubsection{Design for the concept}
Many people engaged not only in the intermediate interaction with the robots but also in the idea and concept behind it. They liked the idea that the robots' behavior would push people to use the trash can more, and the fact that there were two robots encouraged recycling. When they were concerned about the robot, they were concerned about whether the designers were thinking about more vulnerable populations who share the public space, and thinking ahead to the people who might become lazy or have their jobs displaced. Designers of robots must consider how their designs are perceived by the public and what messages are being delivered. As a matter of fact, the first stage of sensemaking was not about the affordances of the robot, but the idea and concept behind it. People's behavior towards the robot largely depends on the result of such a sensemaking process: If people decide that robots have a positive effect on reducing trash, they are likely to welcome and promote them. If people decide the robots are apparatus for surveillance, they will reject, avoid, or even antagonize them, as shown in Clip 3. 

\subsection{Interview Analysis and Video Vignettes for Sensemaking}
In our analysis, we leverage both transcripts from post-interaction interviews and vignettes from the robot's perspective to decipher the sensemaking process behind the interplay between users and robots. Overall, we took the approach where we analyzed transcripts and vignettes individually but linked different resources when possible to analyze the specific instance of interaction further. 

Analyzing interview transcripts gives us insights into both the \textit{sensemaking of deployment} and the \textit{sensemaking of behavior}, where people commented on the concept behind the deployment and the assumptions they made about the robots' functionality. Through thematic analysis, we can observe high-level trends in people's comments. Subsequently, locating video footage within these trends can help us better understand the embodied behaviors of these concepts. For example, people who assume robots are pragmatic are less likely to wave the robot over than people who assume robots are imperative. 

On the other hand, analyzing video vignettes provides us with a first-person perspective of how the interaction unfolds at a turn-taking level. It reinforces some of the themes being discussed in the thematic analysis and expands them with diverse, concrete visual examples. When possible, subsequently analyzing the interview transcripts of users captured in the video provides us with a better understanding of the situated context of the specific interaction from a personal level. 

Overall, the thematic analysis provides a top-down approach to analyzing the high-level patterns that emerged during the sensemaking process. The video vignettes provide a bottom-up opportunity to analyze specific instances in fine detail.

%% file: 06_limitation.tex
\subsection{Limitations}
This research, which focused on one public plaza in one city, is premised on the contextually specific nature of the research. For technical reasons, the wizards did not have access to the robots' onboard video and audio in real-time. Thus, some of the robots' behaviors were not at the human performance level, especially when the participants talked to the robots. Also, since the wizards were not the same on different deployment days, the robots may show different "personalities." While this increases the diversity of robots' interaction styles, the lack of consistency makes categorizing behaviors challenging. Lastly, even though we deployed the robots for two weeks, most participants were first-time users. Thus, the dataset provides little insight into long-term deployments of robots in public spaces. 

In an ideal scenario, we would love to have both video and interview transcripts for every observed interaction. However, due to logistical difficulties, we are only able to interview a portion of the users who used our robot. In addition, the population who agreed to be interviewed were usually fond of the idea of our robots. People who did not like our robot would either leave or refuse to be interviewed. We are aware of the biases that exist in our interview dataset.

%% file: 07_conclusion.tex
\section{Conclusion}

In this work, we analyze video data and interview transcripts from a public deployment of two trash barrel robots in \anon{New York City} to better understand sensemaking activities people perform when they encounter robots in public spaces. We found that people had thoughts not only of the robots or the interactions the robots engaged in, but also the role or the idea of the robots, how other people would respond if they saw the robot, or what the implications on society were. They wanted to understand details about the deployment--are the robots a commercial product? An art project? public infrastructure?--because having that knowledge would change the way they would interact with the robot. Based on our data and analysis, we have provided implications for design that may be topics for future human-robot design researchers who are exploring robots for public space deployment.

%% file: wizard_instruction.tex
For every deployment, two members of the research group (including associated visiting researchers) teleoperated the robots during the deployment as wizards. All of our wizards have backgrounds in computer science and information science. The wizards were encouraged to keep the two robots close to each other during the deployment to contrast their different purposes. Again, different from the constrained instructions given to the wizards in \cite{yang2015experiences}, we took a more explorative approach and allowed for more dynamic control of the robot. 
We provided succinct guidance to the wizards, encouraged them to engage with users in the square in a natural manner, and gave them the freedom to decide how to operate the robot trash barrels based on the situate context.

%% file: interview_questions.tex
These questions were posed to people after they had interacted with a trash barrel robot, after the interviewer got verbal consent both to use the video of the interaction, and to interview them for the study. 

\begin{itemize}
    \item What words come to your mind when you think of the trashbots?
    \item What are things you noticed the trashbots doing that you liked or approved of? Why?
    \item What are things you noticed the trashbots doing that you didn't like or disapproved of? Why?
    \item On a scale from 1 to 10, how do you feel toward the trashbots? (0 means hated it, 10 means loved it.)
    \item Is there anything else you would like to share with us about trashbots?
\end{itemize}

%% file: authorization_consent.tex
The study was conducted with written permission to conduct the experiment in that space from the business improvement district that manages the location, as well as with a certificate of insurance to cover damages that might inadvertently result from the deployment. 

 The protocol was approved under the Cornell University IRB\#806008080; in this protocol, elements of informed consent are waived, based on the IRB finding that the research involves no more than minimal risk to the participants, that the research could not practicably be carried out without the alteration, and that alteration will not adversely affect the rights and welfare of participants (see \cite{common_rule_2018}, \S 46.116(e)(2)). Where people were recorded actively interacting with the robot, we obtained consent post-interaction, and also asked for permission to use images and footage they are featured in. Consent was documented through recorded verbal assent based on \cite{common_rule_2018} \S 46.117(c)(1), as signed consent would be the only record linking the subject and the research, and the principal risk of this minimal-risk protocol would be potential harm resulting from a breach of confidentiality. Consistent with field research conducted in public spaces (see discussion in \cite{sommers2013forgoing}), we did not ask for consent from passersby who were only incidentally involved in the study, even though it could be argued that the use of the unseen "wizard" operators constitutes deception, because those interactions were deemed to be trivial.

%% file: 00_main.bbl

\begin{thebibliography}{48}

%%% the Web address.

\ifx \showCODEN    \undefined \def \showCODEN     #1{\unskip}     \fi
\ifx \showDOI      \undefined \def \showDOI       #1{#1}\fi
\ifx \showISBNx    \undefined \def \showISBNx     #1{\unskip}     \fi
\ifx \showISBNxiii \undefined \def \showISBNxiii  #1{\unskip}     \fi
\ifx \showISSN     \undefined \def \showISSN      #1{\unskip}     \fi
\ifx \showLCCN     \undefined \def \showLCCN      #1{\unskip}     \fi
\ifx \shownote     \undefined \def \shownote      #1{#1}          \fi
\ifx \showarticletitle \undefined \def \showarticletitle #1{#1}   \fi
\ifx \showURL      \undefined \def \showURL       {\relax}        \fi
% invisible to TeX
\providecommand\bibfield[2]{#2}
\providecommand\bibinfo[2]{#2}
\providecommand\natexlab[1]{#1}
\providecommand\showeprint[2][]{arXiv:#2}

\bibitem[Babel et~al\mbox{.}(2022)]%
        {babel2022findings}
\bibfield{author}{\bibinfo{person}{Franziska Babel}, \bibinfo{person}{Johannes Kraus}, {and} \bibinfo{person}{Martin Baumann}.} \bibinfo{year}{2022}\natexlab{}.
\newblock \showarticletitle{Findings from a qualitative field study with an autonomous robot in public: exploration of user reactions and conflicts}.
\newblock \bibinfo{journal}{\emph{International Journal of Social Robotics}} \bibinfo{volume}{14}, \bibinfo{number}{7} (\bibinfo{year}{2022}), \bibinfo{pages}{1625--1655}.
\newblock


\bibitem[Bozikovic(2022)]%
        {Bozikovic_2022}
\bibfield{author}{\bibinfo{person}{Alex Bozikovic}.} \bibinfo{year}{2022}\natexlab{}.
\newblock \showarticletitle{The end of Sidewalk Labs}.
\newblock \bibinfo{journal}{\emph{Architectural Record}} (\bibinfo{date}{Mar} \bibinfo{year}{2022}).
\newblock
\urldef\tempurl%
\url{https://www.architecturalrecord.com/articles/15573-the-end-of-sidewalk-labs}
\showURL{%
\tempurl}


\bibitem[Brown et~al\mbox{.}(2024)]%
        {brown2024trash}
\bibfield{author}{\bibinfo{person}{Barry Brown}, \bibinfo{person}{Fanjun Bu}, \bibinfo{person}{Ilan Mandel}, {and} \bibinfo{person}{Wendy Ju}.} \bibinfo{year}{2024}\natexlab{}.
\newblock \showarticletitle{Trash in Motion: Emergent Interactions with a Robotic Trashcan}. In \bibinfo{booktitle}{\emph{Proceedings of the CHI Conference on Human Factors in Computing Systems}}. \bibinfo{pages}{1--17}.
\newblock


\bibitem[Bu et~al\mbox{.}(2024)]%
        {bu2024field}
\bibfield{author}{\bibinfo{person}{Fanjun Bu}, \bibinfo{person}{Alexandra~WD Bremers}, \bibinfo{person}{Mark Colley}, {and} \bibinfo{person}{Wendy Ju}.} \bibinfo{year}{2024}\natexlab{}.
\newblock \showarticletitle{Field Notes on Deploying Research Robots in Public Spaces}. In \bibinfo{booktitle}{\emph{Extended Abstracts of the CHI Conference on Human Factors in Computing Systems}}. \bibinfo{pages}{1--6}.
\newblock


\bibitem[Bu and Ju(2024)]%
        {bu2024ssup}
\bibfield{author}{\bibinfo{person}{Fanjun Bu} {and} \bibinfo{person}{Wendy Ju}.} \bibinfo{year}{2024}\natexlab{}.
\newblock \showarticletitle{SSUP-HRI: Social Signaling in Urban Public Human-Robot Interaction dataset}.
\newblock \bibinfo{journal}{\emph{arXiv preprint arXiv:2403.10994}} (\bibinfo{year}{2024}).
\newblock


\bibitem[Bu et~al\mbox{.}(2023)]%
        {bu2023trash}
\bibfield{author}{\bibinfo{person}{Fanjun Bu}, \bibinfo{person}{Ilan Mandel}, \bibinfo{person}{Wen-Ying Lee}, {and} \bibinfo{person}{Wendy Ju}.} \bibinfo{year}{2023}\natexlab{}.
\newblock \showarticletitle{Trash barrel robots in the city}. In \bibinfo{booktitle}{\emph{Companion of the 2023 ACM/IEEE International Conference on Human-Robot Interaction}}. \bibinfo{publisher}{ACM/IEEE}, \bibinfo{address}{Stockholm, Sweden}, \bibinfo{pages}{875--877}.
\newblock


\bibitem[Cameron et~al\mbox{.}(2023)]%
        {cameron2023social}
\bibfield{author}{\bibinfo{person}{David Cameron}, \bibinfo{person}{Emily~C Collins}, \bibinfo{person}{Stevienna de Saille}, \bibinfo{person}{Iveta Eimontaite}, \bibinfo{person}{Alice Greenwood}, {and} \bibinfo{person}{James Law}.} \bibinfo{year}{2023}\natexlab{}.
\newblock \showarticletitle{The Social Triad Model: Considering the Deployer in a Novel Approach to Trust in Human--Robot Interaction}.
\newblock \bibinfo{journal}{\emph{International Journal of Social Robotics}} \bibinfo{volume}{0}, \bibinfo{number}{0} (\bibinfo{year}{2023}), \bibinfo{pages}{1--14}.
\newblock
\urldef\tempurl%
\url{https://doi.org/10.1007/s12369-023-01048-3}
\showURL{%
\tempurl}


\bibitem[Clark(1996)]%
        {clark1996using}
\bibfield{author}{\bibinfo{person}{Herbert~H Clark}.} \bibinfo{year}{1996}\natexlab{}.
\newblock \bibinfo{booktitle}{\emph{Using language}}.
\newblock \bibinfo{publisher}{Cambridge University Press}, \bibinfo{address}{Cambridge, UK}.
\newblock


\bibitem[Clark and Fischer(2023)]%
        {clark_fischer_2023}
\bibfield{author}{\bibinfo{person}{Herbert~H. Clark} {and} \bibinfo{person}{Kerstin Fischer}.} \bibinfo{year}{2023}\natexlab{}.
\newblock \showarticletitle{Social robots as depictions of social agents}.
\newblock \bibinfo{journal}{\emph{Behavioral and Brain Sciences}}  \bibinfo{volume}{46} (\bibinfo{year}{2023}), \bibinfo{pages}{e21}.
\newblock
\urldef\tempurl%
\url{https://doi.org/10.1017/S0140525X22000668}
\showDOI{\tempurl}


\bibitem[Cross(2006)]%
        {cross2006designerly}
\bibfield{author}{\bibinfo{person}{Nigel Cross}.} \bibinfo{year}{2006}\natexlab{}.
\newblock \bibinfo{booktitle}{\emph{Designerly ways of knowing}}.
\newblock \bibinfo{publisher}{Springer}, \bibinfo{address}{London, UK}.
\newblock


\bibitem[Dix(2002)]%
        {dix2002beyond}
\bibfield{author}{\bibinfo{person}{Alan Dix}.} \bibinfo{year}{2002}\natexlab{}.
\newblock \showarticletitle{Beyond intention-pushing boundaries with incidental interaction}. In \bibinfo{booktitle}{\emph{Proceedings of Building Bridges: Interdisciplinary Context-Sensitive Computing, Glasgow University}}, Vol.~\bibinfo{volume}{9}. \bibinfo{publisher}{University of Glasgow}, \bibinfo{address}{Glasgow, UK}, \bibinfo{pages}{1--6}.
\newblock
\urldef\tempurl%
\url{https://www.research.lancs.ac.uk/portal/en/publications/beyond-intention--pushing-boundaries-with-incidental-interaction(4c954c9a-0f00-4820-be4c-c92e41631179)/export.html}
\showURL{%
\tempurl}


\bibitem[Fantasia et~al\mbox{.}(2022)]%
        {fantasia2022making}
\bibfield{author}{\bibinfo{person}{Valentina Fantasia}, \bibinfo{person}{Ingar Brinck}, {and} \bibinfo{person}{Christian Balkenius}.} \bibinfo{year}{2022}\natexlab{}.
\newblock \showarticletitle{Making sense with social robots: Extending the landscape of investigation in HRI}. In \bibinfo{booktitle}{\emph{AIC 2022: 8th International Workshop on Artificial Intelligence and Cognition}}. CEUR-WS, \bibinfo{publisher}{CEUR-WS}, \bibinfo{address}{Örebro, Sweden}, \bibinfo{numpages}{8}~pages.
\newblock


\bibitem[Filion et~al\mbox{.}(2023)]%
        {filion2023urban}
\bibfield{author}{\bibinfo{person}{Pierre Filion}, \bibinfo{person}{Markus Moos}, {and} \bibinfo{person}{Gary Sands}.} \bibinfo{year}{2023}\natexlab{}.
\newblock \showarticletitle{Urban neoliberalism, smart city, and Big Tech: The aborted Sidewalk Labs Toronto experiment}.
\newblock \bibinfo{journal}{\emph{Journal of Urban Affairs}} \bibinfo{volume}{0}, \bibinfo{number}{0} (\bibinfo{year}{2023}), \bibinfo{pages}{1--19}.
\newblock
\urldef\tempurl%
\url{https://doi.org/10.1080/07352166.2022.2081171}
\showDOI{\tempurl}


\bibitem[Fischer(2016)]%
        {fischer2016designing}
\bibfield{author}{\bibinfo{person}{Kerstin Fischer}.} \bibinfo{year}{2016}\natexlab{}.
\newblock \bibinfo{booktitle}{\emph{Designing speech for a recipient}}.
\newblock \bibinfo{publisher}{John Benjamins Publishing Company}, \bibinfo{address}{Amsterdam, Netherlands}. 1--337 pages.
\newblock


\bibitem[Fischer et~al\mbox{.}(2021)]%
        {fischer2021same}
\bibfield{author}{\bibinfo{person}{Kerstin Fischer}, \bibinfo{person}{Lars~Christian Jensen}, {and} \bibinfo{person}{Nadine Zitzmann}.} \bibinfo{year}{2021}\natexlab{}.
\newblock \showarticletitle{In the same boat: The Influence of Sharing the Situational Context on a Speaker’s (a Robot’s) Persuasiveness}.
\newblock \bibinfo{journal}{\emph{Interaction Studies}} \bibinfo{volume}{22}, \bibinfo{number}{3} (\bibinfo{year}{2021}), \bibinfo{pages}{488--515}.
\newblock


\bibitem[Fischer et~al\mbox{.}(2015)]%
        {fischer2015initiating}
\bibfield{author}{\bibinfo{person}{Kerstin Fischer}, \bibinfo{person}{Stephen Yang}, \bibinfo{person}{Brian Mok}, \bibinfo{person}{Rohan Maheshwari}, \bibinfo{person}{David Sirkin}, {and} \bibinfo{person}{Wendy Ju}.} \bibinfo{year}{2015}\natexlab{}.
\newblock \showarticletitle{Initiating interactions and negotiating approach: a robotic trash can in the field}. In \bibinfo{booktitle}{\emph{2015 AAAI Spring Symposium Series}}. \bibinfo{publisher}{Association for the Advancement of Artificial Intelligence}, \bibinfo{address}{Stanford}.
\newblock


\bibitem[Forlizzi(2007)]%
        {forlizzi2007robotic}
\bibfield{author}{\bibinfo{person}{Jodi Forlizzi}.} \bibinfo{year}{2007}\natexlab{}.
\newblock \showarticletitle{How robotic products become social products: an ethnographic study of cleaning in the home}. In \bibinfo{booktitle}{\emph{Proceedings of the ACM/IEEE international conference on Human-robot interaction}}. \bibinfo{publisher}{ACM/IEEE}, \bibinfo{address}{Arlington, USA}, \bibinfo{pages}{129--136}.
\newblock


\bibitem[Gretzel and Murphy(2019)]%
        {gretzel2019making}
\bibfield{author}{\bibinfo{person}{Ulrike Gretzel} {and} \bibinfo{person}{Jamie Murphy}.} \bibinfo{year}{2019}\natexlab{}.
\newblock \showarticletitle{Making sense of robots: Consumer discourse on robots in tourism and hospitality service settings}.
\newblock In \bibinfo{booktitle}{\emph{Robots, artificial intelligence, and service automation in travel, tourism and hospitality}}. \bibinfo{publisher}{Emerald Publishing Limited}, \bibinfo{address}{Bingley, England}, \bibinfo{pages}{93--104}.
\newblock


\bibitem[Hallgren(2012)]%
        {hallgren2012computing}
\bibfield{author}{\bibinfo{person}{Kevin~A Hallgren}.} \bibinfo{year}{2012}\natexlab{}.
\newblock \showarticletitle{Computing inter-rater reliability for observational data: an overview and tutorial}.
\newblock \bibinfo{journal}{\emph{Tutorials in quantitative methods for psychology}} \bibinfo{volume}{8}, \bibinfo{number}{1} (\bibinfo{year}{2012}), \bibinfo{pages}{23}.
\newblock


\bibitem[Hoggenmueller et~al\mbox{.}(2020)]%
        {hoggenmueller2020emotional}
\bibfield{author}{\bibinfo{person}{Marius Hoggenmueller}, \bibinfo{person}{Jiahao Chen}, {and} \bibinfo{person}{Luke Hespanhol}.} \bibinfo{year}{2020}\natexlab{}.
\newblock \showarticletitle{Emotional expressions of non-humanoid urban robots: the role of contextual aspects on interpretations}. In \bibinfo{booktitle}{\emph{Proceedings of the 9TH ACM International Symposium on Pervasive Displays}}. \bibinfo{pages}{87--95}.
\newblock


\bibitem[Jung and Hinds(2018)]%
        {jung2018robots}
\bibfield{author}{\bibinfo{person}{Malte Jung} {and} \bibinfo{person}{Pamela Hinds}.} \bibinfo{year}{2018}\natexlab{}.
\newblock \bibinfo{title}{Robots in the wild: A time for more robust theories of human-robot interaction}.
\newblock , \bibinfo{numpages}{5}~pages.
\newblock


\bibitem[Kahn et~al\mbox{.}(2008)]%
        {kahn2008design}
\bibfield{author}{\bibinfo{person}{Peter~H Kahn}, \bibinfo{person}{Nathan~G Freier}, \bibinfo{person}{Takayuki Kanda}, \bibinfo{person}{Hiroshi Ishiguro}, \bibinfo{person}{Jolina~H Ruckert}, \bibinfo{person}{Rachel~L Severson}, {and} \bibinfo{person}{Shaun~K Kane}.} \bibinfo{year}{2008}\natexlab{}.
\newblock \showarticletitle{Design patterns for sociality in human-robot interaction}. In \bibinfo{booktitle}{\emph{Proceedings of the 3rd ACM/IEEE international conference on Human robot interaction}}. \bibinfo{pages}{97--104}.
\newblock


\bibitem[Lee and Sabanovi{\'c}(2014)]%
        {lee2014culturally}
\bibfield{author}{\bibinfo{person}{Hee~Rin Lee} {and} \bibinfo{person}{Selma Sabanovi{\'c}}.} \bibinfo{year}{2014}\natexlab{}.
\newblock \showarticletitle{Culturally variable preferences for robot design and use in South Korea, Turkey, and the United States}. In \bibinfo{booktitle}{\emph{Proceedings of the 2014 ACM/IEEE international conference on Human-robot interaction}}. \bibinfo{publisher}{ACM/IEEE}, \bibinfo{address}{Bielefeld, Germany}, \bibinfo{pages}{17--24}.
\newblock


\bibitem[Lee and Forlizzi(2009)]%
        {lee2009designing}
\bibfield{author}{\bibinfo{person}{Min~Kyung Lee} {and} \bibinfo{person}{Jodi Forlizzi}.} \bibinfo{year}{2009}\natexlab{}.
\newblock \showarticletitle{Designing adaptive robotic services}. In \bibinfo{booktitle}{\emph{Proceedings of IASDR’09}}. \bibinfo{publisher}{IASDR}, \bibinfo{address}{Seoul, Korea}, \bibinfo{pages}{1--10}.
\newblock


\bibitem[Lee et~al\mbox{.}(2019)]%
        {lee2019design}
\bibfield{author}{\bibinfo{person}{Wen-Ying Lee}, \bibinfo{person}{Yoyo Tsung-Yu Hou}, \bibinfo{person}{Cristina Zaga}, {and} \bibinfo{person}{Malte Jung}.} \bibinfo{year}{2019}\natexlab{}.
\newblock \showarticletitle{Design for serendipitous interaction: Bubblebot-bringing people together with bubbles}. In \bibinfo{booktitle}{\emph{2019 14th ACM/IEEE international conference on human-robot interaction (HRI)}}. IEEE, \bibinfo{pages}{759--760}.
\newblock


\bibitem[Lee and Jung(2020)]%
        {lee2020ludic}
\bibfield{author}{\bibinfo{person}{Wen-Ying Lee} {and} \bibinfo{person}{Malte Jung}.} \bibinfo{year}{2020}\natexlab{}.
\newblock \showarticletitle{Ludic-hri: Designing playful experiences with robots}. In \bibinfo{booktitle}{\emph{Companion of the 2020 ACM/IEEE International Conference on Human-Robot Interaction}}. \bibinfo{pages}{582--584}.
\newblock


\bibitem[Lucero(2015)]%
        {lucero2015using}
\bibfield{author}{\bibinfo{person}{Andr{\'e}s Lucero}.} \bibinfo{year}{2015}\natexlab{}.
\newblock \showarticletitle{Using affinity diagrams to evaluate interactive prototypes}. In \bibinfo{booktitle}{\emph{Proceedings of INTERACT 2015: 15th IFIP TC 13 International Conference}}. Springer, \bibinfo{publisher}{Springer}, \bibinfo{address}{Bamberg, Germany}, \bibinfo{pages}{231--248}.
\newblock


\bibitem[Lupetti et~al\mbox{.}(2021)]%
        {lupetti2021designerly}
\bibfield{author}{\bibinfo{person}{Maria~Luce Lupetti}, \bibinfo{person}{Cristina Zaga}, {and} \bibinfo{person}{Nazli Cila}.} \bibinfo{year}{2021}\natexlab{}.
\newblock \showarticletitle{Designerly ways of knowing in HRI: Broadening the scope of design-oriented HRI through the concept of intermediate-level knowledge}. In \bibinfo{booktitle}{\emph{Proceedings of the 2021 ACM/IEEE International Conference on Human-Robot Interaction}}. \bibinfo{publisher}{ACM/IEEE}, \bibinfo{address}{online}, \bibinfo{pages}{389--398}.
\newblock


\bibitem[Lyons et~al\mbox{.}(2023)]%
        {lyons2023explanations}
\bibfield{author}{\bibinfo{person}{Joseph~B Lyons}, \bibinfo{person}{Izz aldin Hamdan}, {and} \bibinfo{person}{Thy~Q Vo}.} \bibinfo{year}{2023}\natexlab{}.
\newblock \showarticletitle{Explanations and trust: What happens to trust when a robot partner does something unexpected?}
\newblock \bibinfo{journal}{\emph{Computers in Human Behavior}}  \bibinfo{volume}{138} (\bibinfo{year}{2023}), \bibinfo{pages}{107473}.
\newblock


\bibitem[Madsbjerg(2019)]%
        {madsbjerg2019sensemaking}
\bibfield{author}{\bibinfo{person}{C. Madsbjerg}.} \bibinfo{year}{2019}\natexlab{}.
\newblock \bibinfo{booktitle}{\emph{Sensemaking: What Makes Human Intelligence Essential in the Age of the Algorithm}}.
\newblock \bibinfo{publisher}{Little, Brown Book Group Limited}, \bibinfo{address}{London}.
\newblock
\showISBNx{9780349142258}


\bibitem[Moshkina et~al\mbox{.}(2014)]%
        {moshkina2014social}
\bibfield{author}{\bibinfo{person}{Lilia Moshkina}, \bibinfo{person}{Susan Trickett}, {and} \bibinfo{person}{J~Gregory Trafton}.} \bibinfo{year}{2014}\natexlab{}.
\newblock \showarticletitle{Social engagement in public places: a tale of one robot}. In \bibinfo{booktitle}{\emph{Proceedings of the 2014 ACM/IEEE international conference on Human-robot interaction}}. \bibinfo{publisher}{ACMIEEE}, \bibinfo{address}{Bielefeld, Germany}, \bibinfo{pages}{382--389}.
\newblock


\bibitem[Papagni and Koeszegi(2020)]%
        {papagni2020understandable}
\bibfield{author}{\bibinfo{person}{Guglielmo Papagni} {and} \bibinfo{person}{Sabine Koeszegi}.} \bibinfo{year}{2020}\natexlab{}.
\newblock \showarticletitle{Understandable and trustworthy explainable robots: A sensemaking perspective}.
\newblock \bibinfo{journal}{\emph{Paladyn, Journal of Behavioral Robotics}} \bibinfo{volume}{12}, \bibinfo{number}{1} (\bibinfo{year}{2020}), \bibinfo{pages}{13--30}.
\newblock


\bibitem[Pelikan et~al\mbox{.}(2024)]%
        {pelikan2024encountering}
\bibfield{author}{\bibinfo{person}{Hannah~RM Pelikan}, \bibinfo{person}{Stuart Reeves}, {and} \bibinfo{person}{Marina~N Cantarutti}.} \bibinfo{year}{2024}\natexlab{}.
\newblock \showarticletitle{Encountering Autonomous Robots on Public Streets}. In \bibinfo{booktitle}{\emph{Proceedings of the 2024 ACM/IEEE International Conference on Human-Robot Interaction}}. \bibinfo{pages}{561--571}.
\newblock


\bibitem[Rudaz et~al\mbox{.}(2023)]%
        {rudaz2023inanimate}
\bibfield{author}{\bibinfo{person}{Damien Rudaz}, \bibinfo{person}{Karen Tatarian}, \bibinfo{person}{Rebecca Stower}, {and} \bibinfo{person}{Christian Licoppe}.} \bibinfo{year}{2023}\natexlab{}.
\newblock \showarticletitle{From inanimate object to agent: Impact of pre-beginnings on the emergence of greetings with a robot}.
\newblock \bibinfo{journal}{\emph{ACM Transactions on Human-Robot Interaction}} \bibinfo{volume}{12}, \bibinfo{number}{3} (\bibinfo{year}{2023}), \bibinfo{pages}{1--31}.
\newblock


\bibitem[Sabanovic et~al\mbox{.}(2006)]%
        {sabanovic2006robots}
\bibfield{author}{\bibinfo{person}{Selma Sabanovic}, \bibinfo{person}{Marek~P Michalowski}, {and} \bibinfo{person}{Reid Simmons}.} \bibinfo{year}{2006}\natexlab{}.
\newblock \showarticletitle{Robots in the wild: Observing human-robot social interaction outside the lab}. In \bibinfo{booktitle}{\emph{9th IEEE International Workshop on Advanced Motion Control, 2006.}} IEEE, \bibinfo{publisher}{IEEE}, \bibinfo{address}{Istanbul, Turkey}, \bibinfo{pages}{596--601}.
\newblock


\bibitem[Salvini(2018)]%
        {salvini2018urban}
\bibfield{author}{\bibinfo{person}{Pericle Salvini}.} \bibinfo{year}{2018}\natexlab{}.
\newblock \showarticletitle{Urban robotics: Towards responsible innovations for our cities}.
\newblock \bibinfo{journal}{\emph{Robotics and Autonomous Systems}}  \bibinfo{volume}{100} (\bibinfo{year}{2018}), \bibinfo{pages}{278--286}.
\newblock


\bibitem[Saupp{\'e} and Mutlu(2015)]%
        {sauppe2015social}
\bibfield{author}{\bibinfo{person}{Allison Saupp{\'e}} {and} \bibinfo{person}{Bilge Mutlu}.} \bibinfo{year}{2015}\natexlab{}.
\newblock \showarticletitle{The social impact of a robot co-worker in industrial settings}. In \bibinfo{booktitle}{\emph{Proceedings of the 33rd annual ACM conference on human factors in computing systems}}. \bibinfo{publisher}{ACM}, \bibinfo{address}{Seoul, Korea}, \bibinfo{pages}{3613--3622}.
\newblock


\bibitem[Siino and Hinds(2005)]%
        {siino2005robots}
\bibfield{author}{\bibinfo{person}{Rosanne~M Siino} {and} \bibinfo{person}{Pamela~J Hinds}.} \bibinfo{year}{2005}\natexlab{}.
\newblock \showarticletitle{Robots, Gender \& amp; Sensemaking: Sex Segregation's Impact On Workers Making Sense Of a Mobile Autonomous Robot}. In \bibinfo{booktitle}{\emph{Proceedings of the 2005 IEEE international conference on robotics and automation}}. IEEE, \bibinfo{publisher}{IEEE}, \bibinfo{address}{Barcelona, Spain}, \bibinfo{pages}{2773--2778}.
\newblock


\bibitem[Sommers and Miller(2013)]%
        {sommers2013forgoing}
\bibfield{author}{\bibinfo{person}{Roseanna Sommers} {and} \bibinfo{person}{Franklin~G Miller}.} \bibinfo{year}{2013}\natexlab{}.
\newblock \showarticletitle{Forgoing debriefing in deceptive research: Is it ever ethical?}
\newblock \bibinfo{journal}{\emph{Ethics \& Behavior}} \bibinfo{volume}{23}, \bibinfo{number}{2} (\bibinfo{year}{2013}), \bibinfo{pages}{98--116}.
\newblock


\bibitem[Szafir and Szafir(2021)]%
        {szafir2021connecting}
\bibfield{author}{\bibinfo{person}{Daniel Szafir} {and} \bibinfo{person}{Danielle~Albers Szafir}.} \bibinfo{year}{2021}\natexlab{}.
\newblock \showarticletitle{Connecting human-robot interaction and data visualization}. In \bibinfo{booktitle}{\emph{Proceedings of the 2021 ACM/IEEE International Conference on Human-Robot Interaction}}. \bibinfo{publisher}{ACM/IEEE}, \bibinfo{address}{online}, \bibinfo{pages}{281--292}.
\newblock


\bibitem[Turner et~al\mbox{.}(2023)]%
        {turner2023multifaceted}
\bibfield{author}{\bibinfo{person}{John~R Turner}, \bibinfo{person}{Jeff Allen}, \bibinfo{person}{Suliman Hawamdeh}, {and} \bibinfo{person}{Gujjula Mastanamma}.} \bibinfo{year}{2023}\natexlab{}.
\newblock \showarticletitle{The multifaceted sensemaking theory: A systematic literature review and content analysis on sensemaking}.
\newblock \bibinfo{journal}{\emph{Systems}} \bibinfo{volume}{11}, \bibinfo{number}{3} (\bibinfo{year}{2023}), \bibinfo{pages}{145}.
\newblock


\bibitem[{U.S. Department of Health and Human Services}(2018)]%
        {common_rule_2018}
\bibfield{author}{\bibinfo{person}{{U.S. Department of Health and Human Services}}.} \bibinfo{year}{2018}\natexlab{}.
\newblock \bibinfo{title}{Code of Federal Regulations, 45 CFR 46, Protection of Human Subjects}.
\newblock \bibinfo{howpublished}{45 CFR 46}.
\newblock
\urldef\tempurl%
\url{https://www.ecfr.gov/cgi-bin/retrieveECFR?gp=&SID=83cd09e1c0f5c6937cd9d7513160fc3f&pitd=20180719&n=pt45.1.46&r=PART&ty=HTML}
\showURL{%
\tempurl}


\bibitem[Weick et~al\mbox{.}(2005)]%
        {weick2005organizing}
\bibfield{author}{\bibinfo{person}{Karl~E Weick}, \bibinfo{person}{Kathleen~M Sutcliffe}, {and} \bibinfo{person}{David Obstfeld}.} \bibinfo{year}{2005}\natexlab{}.
\newblock \showarticletitle{Organizing and the process of sensemaking}.
\newblock \bibinfo{journal}{\emph{Organization science}} \bibinfo{volume}{16}, \bibinfo{number}{4} (\bibinfo{year}{2005}), \bibinfo{pages}{409--421}.
\newblock


\bibitem[Weinberg et~al\mbox{.}(2023)]%
        {weinberg2023sharing}
\bibfield{author}{\bibinfo{person}{David Weinberg}, \bibinfo{person}{Healy Dwyer}, \bibinfo{person}{Sarah~E Fox}, {and} \bibinfo{person}{Nikolas Martelaro}.} \bibinfo{year}{2023}\natexlab{}.
\newblock \showarticletitle{Sharing the Sidewalk: Observing Delivery Robot Interactions with Pedestrians during a Pilot in Pittsburgh, PA}.
\newblock \bibinfo{journal}{\emph{Multimodal Technologies and Interaction}} \bibinfo{volume}{7}, \bibinfo{number}{5} (\bibinfo{year}{2023}), \bibinfo{pages}{53}.
\newblock


\bibitem[Weiss et~al\mbox{.}(2008)]%
        {weiss2008methodological}
\bibfield{author}{\bibinfo{person}{Astrid Weiss}, \bibinfo{person}{Regina Bernhaupt}, \bibinfo{person}{Manfred Tscheligi}, \bibinfo{person}{Dirk Wollherr}, \bibinfo{person}{Kolja Kuhnlenz}, {and} \bibinfo{person}{Martin Buss}.} \bibinfo{year}{2008}\natexlab{}.
\newblock \showarticletitle{A methodological variation for acceptance evaluation of human-robot interaction in public places}. In \bibinfo{booktitle}{\emph{RO-MAN 2008-The 17th IEEE International Symposium on Robot and Human Interactive Communication}}. IEEE, \bibinfo{publisher}{IEEE}, \bibinfo{address}{Munich, Germany}, \bibinfo{pages}{713--718}.
\newblock


\bibitem[While et~al\mbox{.}(2021)]%
        {while2021urban}
\bibfield{author}{\bibinfo{person}{Aidan~H While}, \bibinfo{person}{Simon Marvin}, {and} \bibinfo{person}{Mateja Kovacic}.} \bibinfo{year}{2021}\natexlab{}.
\newblock \showarticletitle{Urban robotic experimentation: San Francisco, Tokyo and Dubai}.
\newblock \bibinfo{journal}{\emph{Urban Studies}} \bibinfo{volume}{58}, \bibinfo{number}{4} (\bibinfo{year}{2021}), \bibinfo{pages}{769--786}.
\newblock


\bibitem[Yang et~al\mbox{.}(2015)]%
        {yang2015experiences}
\bibfield{author}{\bibinfo{person}{Stephen Yang}, \bibinfo{person}{Brian Ka-Jun Mok}, \bibinfo{person}{David Sirkin}, \bibinfo{person}{Hillary~Page Ive}, \bibinfo{person}{Rohan Maheshwari}, \bibinfo{person}{Kerstin Fischer}, {and} \bibinfo{person}{Wendy Ju}.} \bibinfo{year}{2015}\natexlab{}.
\newblock \showarticletitle{Experiences developing socially acceptable interactions for a robotic trash barrel}. In \bibinfo{booktitle}{\emph{2015 24th IEEE International Symposium on Robot and Human Interactive Communication (RO-MAN)}}. IEEE, \bibinfo{publisher}{IEEE}, \bibinfo{address}{Portland, USA}, \bibinfo{pages}{277--284}.
\newblock


\bibitem[Zamfirescu-Pereira et~al\mbox{.}(2021)]%
        {fakeittomakeit}
\bibfield{author}{\bibinfo{person}{J.D. Zamfirescu-Pereira}, \bibinfo{person}{David Sirkin}, \bibinfo{person}{David Goedicke}, \bibinfo{person}{Ray LC}, \bibinfo{person}{Natalie Friedman}, \bibinfo{person}{Ilan Mandel}, \bibinfo{person}{Nikolas Martelaro}, {and} \bibinfo{person}{Wendy Ju}.} \bibinfo{year}{2021}\natexlab{}.
\newblock \showarticletitle{Fake It to Make It: Exploratory Prototyping in HRI}. In \bibinfo{booktitle}{\emph{Companion of the 2021 ACM/IEEE International Conference on Human-Robot Interaction}} (Boulder, CO, USA) \emph{(\bibinfo{series}{HRI '21 Companion})}. \bibinfo{publisher}{Association for Computing Machinery}, \bibinfo{address}{New York, NY, USA}, \bibinfo{pages}{19–28}.
\newblock
\showISBNx{9781450382908}
\urldef\tempurl%
\url{https://doi.org/10.1145/3434074.3446909}
\showDOI{\tempurl}


\end{thebibliography}
